\documentclass[11pt,a4paper]{article}

\usepackage[a4paper,margin=2.2cm]{geometry}

\usepackage{graphicx}
\usepackage{amsmath,amsfonts,amssymb}
\allowdisplaybreaks

\usepackage[authoryear,longnamesfirst]{natbib}

\usepackage{algorithm}
\usepackage{algpseudocode}
\usepackage{caption}
\usepackage{subfig}
\usepackage{multirow}
\usepackage{booktabs}
\usepackage{array}
\usepackage{float}
\usepackage{placeins}

\usepackage{xcolor}
\usepackage{url}
\usepackage[hidelinks]{hyperref}

\usepackage{xspace}
\usepackage{etoolbox}
\def\tsc#1{\csdef{#1}{\textsc{\lowercase{#1}}\xspace}}
\tsc{WGM}
\tsc{QE}

\begin{document}

\title{Quantum Annealing Enhanced Reinforcement Learning for Accurate Remaining Useful Lifetime Prediction}

\author{
  Manoranjan Gandhudi\textsuperscript{a} \quad
  Arunkumar V.\textsuperscript{b} \quad
  G.R. Anil\textsuperscript{c} \quad
  Gangadharan G.R.\textsuperscript{d}
}
\date{}

\maketitle

\begin{center}\footnotesize
\textsuperscript{a}Central University of Karnataka, Kalaburagi, India, \href{mailto:gmanoranjan@cuk.ac.in}{gmanoranjan@cuk.ac.in};\quad
\textsuperscript{b}University College of Engineering, Anna University, Tiruchirappalli, Tamil Nadu, India, \href{mailto:arunkumarv1530@gmail.com}{arunkumarv1530@gmail.com};\quad
\textsuperscript{c}AIONOS India Pvt Ltd, Hyderabad, India, \href{mailto:anilgrcse@gmail.com}{anilgrcse@gmail.com};\quad
\textsuperscript{d}National Institute of Technology Tiruchirappalli, India, \href{mailto:ganga@nitt.edu}{ganga@nitt.edu}
\end{center}

\begin{abstract}
Remaining useful life estimation is central to predictive maintenance and reliability engineering, where the cost of an unplanned failure on an aircraft engine or a connected industrial device can dwarf the cost of the asset itself. Traditional statistical degradation models fail to capture the strongly nonlinear behavior of real systems. Data-driven machine learning models have improved accuracy, but they converge to suboptimal solutions in high-dimensional and non-convex search spaces. To address these limitations, this study proposes a quantum annealing enhanced Q-Learning framework that couples the sampling behaviour of quantum annealing with the sequential decision making structure of Q-learning. Each Q-value update is encoded as a small quadratic unconstrained binary optimization whose ground state is the greedy action under the current value estimates; rather than acting as a deterministic optimizer, the annealer returns a distribution over near-optimal actions across many reads, and this stochastic action selection provides exploration that helps the agent avoid premature convergence on highly nonlinear degradation trajectories. The QUBO is solved on the D-Wave Advantage system using minor embedding, with an annealing time of 20\,$\mu$s and 1{,}000 reads per Q-value update. The framework is validated on two public benchmarks, including the NASA C-MAPSS turbofan engine and the device-fleet predictive maintenance datasets. Averaged over 30 independent runs, QAQL attains a MSE of 435.28 $\pm$ 12.4, 593.69 $\pm$ 50.22, 549.54 $\pm$ 14.24, and 880.59 $\pm$ 260.68 on C-MAPSS FD001, FD002, FD003, and FD004 respectively and a MSE of 126.28 $\pm$ 4.1 on the predictive maintenance dataset, outperforming the classical and quantum baselines considered in this study across six error metrics. The results indicate that quantum annealing is a usable, not merely theoretical, optimizer inside a reinforcement-learning loop for industrial applications.
\end{abstract}

\medskip
\noindent\textbf{Highlights}
\begin{itemize}
\item Quantum-annealing-enhanced Q-learning (QAQL) is proposed for RUL prediction.
\item The greedy action step is recast as a QUBO and sampled on the D-Wave Advantage QPU.
\item Annealer sampling adds exploration that curbs the premature convergence of classical RL on nonlinear degradation.
\item Outperforms the 14 classical and quantum baselines evaluated under a common protocol on six error metrics ($p<0.01$).
\end{itemize}


\vspace{1em}

\section{Introduction}
\label{sec:introduction}
A turbofan engine on a boeing 767 ingested a fan blade that had been undergoing undetected fatigue cracking for approximately 1{,}200 flight cycles \citep{yang2025failure}. This incident highlights the critical importance of predictive maintenance and early fault detection in aerospace systems to prevent catastrophic engine failures and ensure operational safety \citep{wang2025online}. The aircraft landed safely, but the unscheduled removal cost the operator close to USD 1.4 million in lost revenue and repair, and grounded the airframe for eleven days. Industrial reports place the global cost of unplanned downtime at over USD 1.5 trillion per year \citep{magadan2024robust, liu2024optimized}, and the single most effective lever for reducing that figure is an accurate estimate of how long a component can keep operating before it needs to come off the wing, off the production line, or off the grid. That quantity is called the Remaining Useful Lifetime (RUL), and the gap between a good estimate and a poor one is the gap between predictive maintenance and reactive maintenance.

Early RUL estimators were physics-driven \citep{tang2026physics}. They used Wiener processes, gamma processes, and Paris-Erdogan crack-growth laws to fit closed-form degradation curves to a small number of monitored variables. These models are interpretable, but they break down on modern assets that are instrumented with dozens of correlated sensors, that operate over multiple flight regimes or load cycles, and whose degradation paths are non-monotonic. The community responded by moving to data-driven methods. Convolutional, recurrent, and attention-based deep networks now dominate the public RUL benchmarks \citep{ferreira2022remaining, magadan2024robust}, and on the NASA C-MAPSS turbofan dataset they have pushed the FD001 RMSE down from above 30 cycles in 2014 to roughly 11 cycles today. Two limitations still hold the field back. First, deep models are hungry for labelled run-to-failure trajectories, which are rare in practice because most fielded equipment is taken out of service before it actually fails. Second, the loss surfaces involved are highly non-convex, and gradient-based optimisers routinely converge to policies that fit the average degradation pattern but miss the long tail of unusual failure modes that matter most operationally \citep{ferreira2022remaining}.

A complementary line of work treats RUL as a sequential decision problem and uses reinforcement learning (RL) to learn a maintenance-aware value function directly from sensor histories \citep{shakya2023reinforcement, 10070849}. Q-learning is the natural starting point because it is model-free and updates a tabular or parametric Q(s,a) function from one-step transitions \citep{evangelidis2024efficient}. The catch is well known: in the high-dimensional, non-stationary state spaces of industrial degradation data, classical Q-learning suffers from slow convergence, high variance, and a strong tendency to settle into locally optimal policies that the usual epsilon-greedy schedule explores away from only slowly. The bottleneck is therefore the exploration that drives the value update, not the arithmetic of the update itself.

Although RUL estimation can be formulated as a supervised regression problem, practical predictive maintenance involves sequential observations collected along degradation trajectories rather than independent samples. In such settings, reinforcement learning provides two advantages. First, temporal-difference learning propagates information backward through the degradation trajectory, allowing future degradation evidence to influence earlier health-state estimates. Second, RL naturally accommodates maintenance-oriented reward functions that can incorporate asymmetric costs associated with early and late predictions. Therefore, RL is employed not because supervised learning is incapable of predicting RUL, but because it provides a trajectory-aware learning framework that can be extended naturally toward maintenance decision optimization.

Quantum annealing offers a different way to drive that step. Programmable annealers such as the D-Wave Advantage system expose 5{,}640 superconducting qubits arranged in a Pegasus topology and solve a Quadratic Unconstrained Binary Optimisation (QUBO) problem by adiabatically evolving the system Hamiltonian from a transverse-field mixer to a problem Hamiltonian whose ground state encodes the optimum \citep{yarkoni2022quantum, khan2020machine}. Crucially for our purpose, a single anneal is not a deterministic solver: each of the many reads returns a sample, and quantum tunnelling and superposition shape the distribution from which those samples are drawn. The question this paper asks is whether that sampling behaviour, rather than any one-shot optimality guarantee, can usefully shape the repeated action selection that sits at the heart of an RL update.

Our hypothesis is that it does, provided the temporal-difference (TD) target is encoded as a QUBO that the annealer can actually solve. We propose Quantum Annealing enhanced Q-Learning (QAQL), a hybrid framework in which the greedy action selection inside each Q-value update is reformulated as a small QUBO whose ground state is the highest-value action under the current Q-table. Because the annealer is queried with many reads, it returns a distribution over near-optimal actions rather than a single deterministic choice; this sampling is what couples quantum annealing to the learning dynamics and supplies the exploration that the update consumes. The selected action is decoded from the lowest-energy sample, applied to the classical Q-table through the standard TD rule, and used to emit the RUL prediction. The annealer is therefore not a generic black-box optimiser bolted on after training; it is woven into the RL loop and sees a problem-specific Hamiltonian on every step.

The objective of quantum annealing in QAQL is not computational acceleration of a small argmax operation. Instead, the annealer acts as a stochastic low-energy sampler integrated within the RL update process. Unlike deterministic greedy selection, the annealer returns a distribution of near-optimal actions across multiple reads. This sampling behaviour increases exploration diversity during value updates and reduces premature convergence to locally optimal policies. The contribution of QAQL therefore lies in exploiting quantum-generated sampling dynamics rather than claiming a computational speedup over classical optimization.

\textbf{Research questions.}
\begin{itemize}
    \item RQ1: Can a per-step QUBO encoding of the TD update let quantum annealing replace the classical Q-value optimisation step inside Q-learning, and does this improve RUL prediction accuracy?
    \item RQ2: Does the resulting hybrid converge faster and to a better policy than classical RL in the high-dimensional, non-convex degradation environments that characterise real industrial data?
    \item RQ3: How robust is QAQL across operating regimes and degradation patterns, and how does it compare to recent classical and quantum baselines on standard benchmarks?
\end{itemize}

\textbf{Research objectives.}
\begin{itemize}
    \item Design a Q-learning update in which the TD-error minimisation is recast as a QUBO compatible with current quantum-annealing hardware.
    \item Implement the resulting QAQL framework end-to-end on the D-Wave Advantage QPU and integrate it with classical preprocessing for industrial sensor data.
    \item Evaluate QAQL against a balanced set of seven classical and seven quantum RUL baselines on two public benchmarks, with statistical-significance testing and a runtime comparison.
\end{itemize}

\textbf{Research outcomes.}
\begin{itemize}
    \item Encoding the greedy action step as a QUBO lets quantum annealing inject sampling-based exploration into the value update, improving RUL accuracy over classical Q-learning on both benchmarks (RQ1).
    \item QAQL converges in fewer episodes and produces lower-variance policies than classical RL baselines, consistent with the broader exploration that annealer sampling provides (RQ2).
    \item Across thirty independent runs and six error metrics, QAQL outperforms the classical and quantum baselines considered in this study, with statistically significant improvements ($p < 0.01$, paired Wilcoxon signed-rank) (RQ3).
\end{itemize}

The remainder of the paper is organised as follows. Section~\ref{sec22} surveys the related work on classical, RL-based, and quantum approaches to RUL prediction. Section~\ref{sec2} formalises the QAQL framework, including the QUBO encoding of the TD update and the integration with the D-Wave Advantage QPU. Section~\ref{sec3} reports experiments on the C-MAPSS FD001 and Predictive Maintenance datasets, with ablations, runtime measurements, and significance tests. Section~\ref{sec4} concludes and outlines extensions to gate-model quantum RL.

\section{Literature Survey}\label{sec22}
\citet{jiao2023lightgbm} integrated a LightGBM architecture integrated with electrochemical modeling for estimating the RUL of batteries under various driving conditions. However, the limitation of the model lies in the need for extensive data preprocessing, which limit its practicality in real time applications. \citet{tian2023novel} proposed a technique that employs divergence of Kullback–Leibler to assess data distribution differences that effectively leverages information from the source to increase the generalization for unseen target disciplines. However, the framework lacks effective feature engineering and does not incorporate a hybrid model combined with physical mechanisms. \citet{hou2026lvdacnn} proposed a lightweight variable dependency aware convolutional neural network (LVDACNN) for RUL prediction. The model enhances variable relationship learning through an embedding scheme and leverages channel attention to suppress irrelevant information by achieving accurate predictions with reduced computational cost. However, its performance may degrade under highly nonlinear degradation patterns and requires careful design of variable dependency representations. 

\citet{wilberforce2023remaining} developed an integrated convolutional bi-RNN for estimating the RUL of fuel cells. However the effectiveness of the model is very limited. The ML and DL models train on huge amounts of labeled data. Potential overfitting challenges in interpreting the predictions and struggle with generalizing to unseen conditions. These will make them less reliable in real world environments where operating conditions can vary dynamically. \citet{xu2023global} implemented an attention approach for accurate RUL predictions that eliminates the use of RNN or CNN modules. The model identifies relevant data features by addressing existing challenges. However, it lacks to assess the impact of feature selection and unable to explore information aggregation within global attention. \citet{pan2022transfer} developed a fusion model combining transfer learning based LSTM and Particle Filter (PF) for lithium ion battery RUL estimation. The model enhances generalization across varying operating conditions and PF provides uncertainty quantification beyond single-point estimates. However, the hybrid framework increases model complexity and computational burden and its performance depends on the quality of source domain training data. \citet{sun2022remaining} implemented a RUL prediction approach for AC contactors using hybrid dual attention LSTM network. The model extracts degradation features and detects inflection points using the feature and temporal attention mechanisms to enhance prediction accuracy. The approach lacks to generalize the challenges under varying operating conditions and relies on handcrafted feature extraction. 

\citet{zhang2023data} proposed an interactive RUL estimation framework integrating PF temporal attention with bi-GRU for analyzing the lithium ion batteries. The approach integrated to capture time dependent significance and degradation uncertainty. It enhances online prediction performance by amalgamating the model based and data driven strategies. However, the model requires accurate parameter tuning for stable deployment. \citet{wang2023comprehensive} developed a graph based network approach for aircraft RUL prediction. The model integrates dynamic graph learning to capture time varying relationships in condition monitoring data while incorporating aero engine structural characteristics. The complexity and graph construction process of the model fails to optimize the computational cost. \citet{liu2026remaining} proposed framework for bearing RUL prediction for cross domain conditions. This employs denoising autoencoders for feature augmentation and an interpretable attention module to highlight degradation patterns. The model relies on source target similarity and increases complexity due to multiple interconnected modules. 

\citet{hu2021reinforcement} proposed an RL strategy with a combined aircraft maintenance simulation. The strategy focuses on a maintenance object to necessitate an extension to accommodate a numerous maintenance scenarios for enhanced applicability. \citet{peng2023health} proposed a RL method to construct health indicators using multisensor data, directly linking health indicators construction with the RUL prediction. However, the method requires improvements in the reliability and accuracy of the predicted results. \citet{abbas2024hierarchical} proposed a model fusing RL and probabilistic model to maximize interpretability in safety of predictive maintenance. The model activates in specific circumstances recognized by an input and output hidden markov decision model, particularly in crucial or near failure conditions. However, it does not fully address abnormal conditions, which limits its effectiveness. \citet{cao2023stochastic} proposed a framework for RUL prediction that addresses aleatory and epistemic uncertainties by combining uncertainty and probability theories to create a stochastic uncertain degradation model. However, further attention is required on prediction methods for nonlinear degradation models, and uncertainty aspects require further exploration. \citet{landau2026federated} implemented a federated learning framework for aircraft RUL estimation by enabling multiple airlines to jointly train a prognostic model. A decentralized validation and novel parameter aggregation methods can enhance robustness against noisy sensor data. However, data heterogeneity and convergence instability remain challenges in large scale federated deployment.

\citet{wang2023dynamic} proposed an unsupervised quantum clustering model used to identify normal and abnormal batteries, enhanced by a Weibull wave function for greater sensitivity to abnormal features. Adjusting the threshold results in either misjudgement of normal batteries or insufficient detection of abnormal batteries, highlighting the need for further optimisation. \citet{ghosh2023evolving} presented a quantum-driven fuzzy model that predicts battery capacity fade under incomplete discharge conditions. However, the model does not account for temperature variation, an important factor in battery health indicators, due to limitations of testing conditions. \citet{tsurkan2025hybrid} developed a hybrid quantum RNN for jet-engine RUL prediction. The network integrates classical dense layers with quantum LSTM layers on the C-MAPSS dataset. However, its scalability is constrained by circuit depth and qubit limitations, increasing training complexity and restricting real-world deployment. \citet{gandhudi2026dynamic} proposed a QNN for RUL prediction optimised using QA, addressing hyperparameter tuning and nonlinear degradation by improving convergence and generalisation. The performance is sensitive to problem encoding and parameter initialisation, which may affect reproducibility and stability across diverse real-world conditions. The related work is summarised in Table~\ref{tab1}.

\begin{table*}[htbp]
\centering
\caption{Summary of Existing RUL Prediction Approaches}
\label{tab1}
\begin{tabular}{p{2.7cm}p{3.2cm}p{4.1cm}p{4.5cm}}
\hline
\textbf{Reference} & \textbf{Model} & \textbf{Objective} & \textbf{Limitations} \\
\hline
\citet{hu2021reinforcement} 
& Q-learning based RL 
& Aircraft maintenance optimization 
& Limited to specific maintenance objects \\

\citet{sun2022remaining} 
& MMPE + Dual-Attention LSTM 
& AC contactor RUL prediction with feature extraction 
& Relies on handcrafted features; limited generalization \\

\citet{jiao2023lightgbm} 
& LightGBM + Electrochemical Model 
& Battery RUL estimation under varying driving conditions 
& Requires extensive preprocessing; limited real-time practicality \\

\citet{cao2023stochastic} 
& Stochastic Uncertain Degradation Model 
& Address aleatory and epistemic uncertainties in RUL prediction 
& Limited handling of nonlinear degradation; uncertainty modeling needs further exploration \\

\citet{xu2023global} 
& Global Attention Mechanism 
& Accurate RUL prediction without CNN/RNN modules 
& Limited analysis of feature selection and information aggregation \\

\citet{peng2023health} 
& RL-based Health Indicator Construction 
& Construct health indicators from multisensor data 
& Needs improvement in reliability and prediction accuracy \\

\citet{zhang2023data} 
& PF + BiGRU + Temporal Attention 
& Lithium-ion battery RUL with uncertainty modeling 
& High computational complexity; sensitive parameter tuning \\

\citet{wang2023comprehensive} 
& Comprehensive Dynamic Structure GNN (CDSG) 
& Aircraft RUL prediction with dynamic graph learning 
& Complex graph construction; high computational cost \\

\citet{tian2023novel} 
& Metric Learning + KL Divergence 
& Improve cross-domain generalization for unseen target domains 
& Lacks hybrid physical modeling and effective feature engineering \\

\citet{wang2023dynamic} 
& Unsupervised Quantum Clustering + Weibull Wave Function 
& Identify normal and abnormal batteries 
& Threshold sensitivity causes misclassification issues \\

\citet{ghosh2023evolving} 
& Quantum Fuzzy Neural Network 
& Battery capacity fade prediction under incomplete discharge 
& Does not consider temperature variations \\

\citet{abbas2024hierarchical} 
& RL + Probabilistic Model (HMDP-based) 
& Improve interpretability in predictive maintenance 
& Does not fully address abnormal conditions \\

\citet{tsurkan2025hybrid} 
& Hybrid Quantum RNN (Quantum LSTM + Classical Layers) 
& Jet engine RUL prediction on C-MAPSS 
& Limited scalability due to qubit and circuit depth constraints \\

\citet{hou2026lvdacnn} 
& Lightweight Variable Dependency Aware CNN (LVDACNN) 
& Efficient RUL prediction with reduced computational cost 
& Performance degrades under highly nonlinear degradation \\

\citet{landau2026federated} 
& Federated Learning Framework 
& Collaborative aircraft engine RUL prediction without data sharing 
& Communication overhead; data heterogeneity; convergence instability \\

\citet{liu2026remaining} 
& STIC (Transfer Learning + Interpretable Attention) 
& Small-sample cross-domain bearing RUL prediction 
& Dependent on source-target similarity; increased training complexity \\

\citet{gandhudi2026dynamic} 
& Quantum Neural Network + Quantum Annealing 
& Optimize nonlinear RUL prediction and hyperparameter tuning 
& Sensitive to encoding and parameter initialization \\

\hline
\end{tabular}
\end{table*}

Three patterns recur across this body of work. First, classical and deep models depend on handcrafted features or extensive preprocessing that limit their real-time applicability and constrain transfer across systems. Second, RL-based prognostic methods inherit the well-known optimisation pathologies of classical Q-learning: slow convergence, high variance, and sensitivity to local minima in the action-value landscape. Third, existing quantum approaches treat quantum computation either as a feature extractor (quantum kernels, QLSTM) or as a hyperparameter optimiser run once outside the learning loop. None of them place a quantum solver inside the per-step value-update of an RL agent. QAQL is positioned at this gap: it keeps the lightweight Q-learning skeleton, but expresses the inner greedy action selection as a QUBO that the D-Wave Advantage QPU samples on every step, so that quantum annealing acts inside the per-step value-update itself. The per-step selection is intentionally small and easy to encode; the contribution is not that annealing solves a hard combinatorial problem at each step, but that querying the QPU repeatedly inside the loop is feasible on current hardware and that its sampling supplies exploration which a deterministic argmax does not.

\section{QAQL: Quantum Annealing Enhanced Q-Learning} \label{sec2}

Remaining Useful Life (RUL) estimation is conventionally formulated as a supervised regression problem in which a model learns a mapping from sensor observations to the corresponding RUL value by minimizing prediction error. From a purely predictive perspective, supervised learning provides a natural and effective framework for this task. Therefore, we do not claim that reinforcement learning (RL) is a necessary replacement for supervised regression. Instead, we investigate whether a trajectory-aware RL framework can provide complementary advantages in degradation modelling and predictive maintenance applications.

First, unlike conventional regression models that optimize prediction accuracy independently for each sample, RL learns from sequential degradation trajectories through temporal-difference (TD) learning. The TD mechanism propagates information obtained from future observations back to earlier health states, enabling the learning process to exploit long-term degradation patterns rather than relying solely on instantaneous prediction errors. This property is particularly relevant in prognostics, where degradation evolves continuously over time and the health state at a given instant is strongly influenced by preceding and future operating conditions.

Second, RL naturally supports reward formulations that incorporate operational and maintenance objectives beyond prediction accuracy. In practical maintenance settings, the consequences of overestimating and underestimating RUL are often asymmetric. Overly optimistic predictions may result in unexpected failures, whereas overly conservative predictions may lead to premature maintenance and increased operational costs. By defining suitable reward functions, RL can explicitly account for such cost-sensitive considerations during learning, providing a flexible framework that extends beyond conventional error minimization.

Third, the RL formulation establishes a natural pathway toward future maintenance decision-making systems. The present study considers an exogenous-transition setting in which actions correspond to discretized RUL estimates and the objective is accurate prognostic prediction. However, industrial predictive maintenance systems increasingly require decision support capabilities in addition to prognosis. In future extensions, the same framework can be adapted such that actions represent maintenance decisions, such as continuing operation, scheduling an inspection, performing a repair, or replacing a component. In such settings, system transitions become action dependent and the problem evolves from a prediction task into a genuine sequential decision-making problem, for which reinforcement learning provides a principled optimization framework. Consequently, the proposed approach may be viewed as an intermediate step toward integrated prognostics and maintenance decision optimization.

\begin{figure}
\centering
    \includegraphics[width=\linewidth]{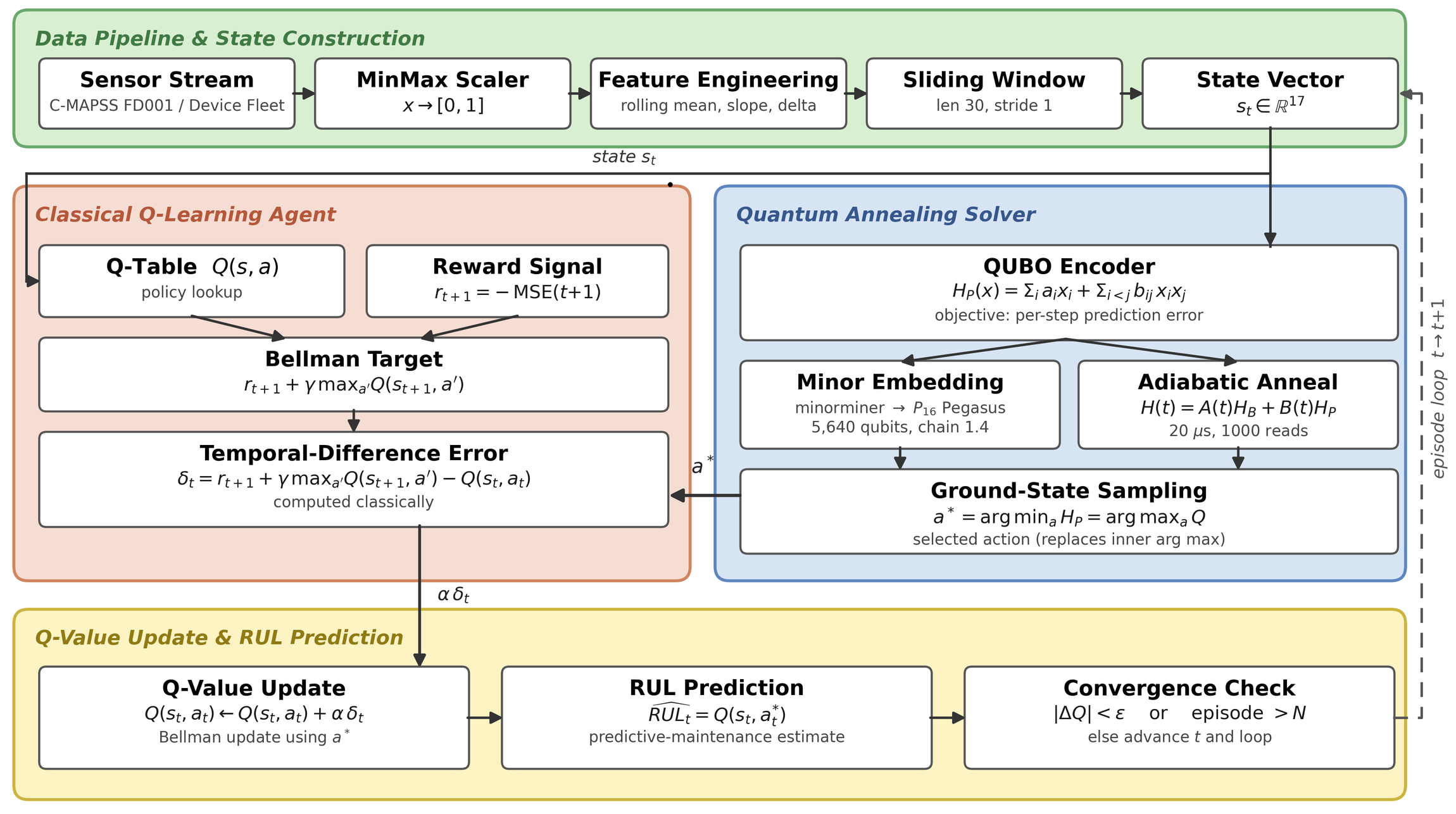}
    \caption{Framework for Proposed Methodology.}
    \label{fig1}
\end{figure}

The proposed framework departs from prior work in one specific way: rather than using quantum annealing as a general-purpose hyperparameter optimiser run once outside the learning loop, QAQL invokes the annealer on every step to solve a small, problem-specific QUBO that encodes the temporal-difference error. The TD error itself is constructed so that minimising it is equivalent to reducing the per-step RUL prediction error, which keeps the quantum solver aligned with the engineering objective at all times. We first introduce the symbols used throughout the paper (Table~\ref{tab333}), then build the framework up from classical Q-learning to the QUBO encoding to the full algorithm (Figure~\ref{fig1}).

\begin{table}[ht]
\centering
    \caption{Symbol Table}
    \begin{tabular}{p{2cm}p{7.5cm}}
    \hline
    \textbf{Symbol} & \textbf{Description} \\
    \hline
    s & State \\
    a & Action \\
    r & Reward \\
    Q(s,a) & Expected reward for action $a$ in state $s$ \\
    $\alpha$ & Learning rate \\
    $\gamma$ & Discount factor \\
    $s'$ & Next state\\
    QUBO & Quadratic Unconstrained Binary Optimization \\
    H & Hamiltonian \\
    $\sigma_{i}^{x}$ & Pauli-X matrices representing quantum superposition. \\
    $|\psi(T)\rangle$ & Quantum state at time $T$. \\
    TD & Temporal Difference\\
    $m$ & Number of discretised RUL bins, $m = |A|$ \\
    $x_i$ & Binary indicator that bin $a_i$ is selected ($x_i \in \{0,1\}$) \\
    $\rho_j$ & Representative value (centre) of bin $a_j$ \\
    $c_j(s)$ & Per-bin selection cost $-Q(s,a_j)$ (negative learned action value) \\
    $\lambda$ & One-hot penalty weight in the QUBO \\
    $a_i, b_{ij}$ & Linear and quadratic QUBO coefficients \\
    $h_i, J_{ij}$ & Linear and quadratic Ising couplings \\
    \hline
    \end{tabular}
    \label{tab333}
\end{table}

Q-learning is a model-free RL algorithm in which an agent learns an action-value function $Q(s,a)$ from one-step transitions, without explicit knowledge of the environment dynamics~\citep{jin2024feasible, evangelidis2024efficient}. In our setting:
\begin{itemize}
    \item State $s$: a vector of normalised sensor readings and engineered features describing the current health condition of the asset.
    \item Action $a$: a discretised RUL prediction drawn from a fixed bin set covering the asset's plausible remaining-life range.
    \item Reward $r$: a function of the prediction accuracy at the current step, defined to penalise large RUL errors.
    \item Q-value $Q(s,a)$: the expected cumulative reward of predicting RUL bin $a$ in state $s$ and following the current policy thereafter.
\end{itemize}

Formally, one run-to-failure trajectory is an episodic, finite-horizon Markov decision process $\mathcal{M} = (S, A, P, R, \gamma)$~\citep{sutton2018reinforcement}. The state $s_t \in S$ is the length-$w$ sliding window of scaled sensor and engineered features ending at cycle $t$; the action $a_t \in A$ is the predicted RUL bin; the reward $R(s_t, a_t) = -\text{MSE}(s_t, a_t)$ penalises the squared prediction error; and $\gamma \in [0,1)$ discounts future error. The transition kernel $P(s_{t+1} \mid s_t, a_t)$ is governed by the asset's physical degradation: the window advances one cycle along the trajectory, and the episode terminates at the failure cycle. Because the next health state is driven by degradation physics rather than by the agent's prediction, the transitions are \emph{exogenous} with respect to the action. This is the established prognostics-as-RL setting adopted in prior RUL work~\citep{skolik2022quantum, peng2023health, hu2021reinforcement}, and we make the assumption explicit here because it determines what the learning signal must do.

The classical Q-learning update is given in Equation~\ref{eq1}.
\begin{equation}\label{eq1}
    Q(s,a) \leftarrow Q(s,a) + \alpha \left[ r + \gamma \max_{a'} Q(s',a') - Q(s,a) \right]
\end{equation}
where $\alpha$ is the learning rate, $\gamma$ is the discount factor, $s'$ is the next state, and $\max_{a'} Q(s',a')$ is the bootstrapped value of the best successor action. The bracketed term is the temporal-difference (TD) error and is the only learning signal the agent receives. The optimisation problem solved by the inner $\arg\max$ becomes the bottleneck in high-dimensional state-action spaces, which is what we replace with a quantum-annealing call.

\textbf{Relationship to supervised regression, stated precisely.} Because the transitions are exogenous, the action cannot influence the next state, and we are explicit about what this implies. In that case the optimal action value factorises as
\begin{equation}\label{eq:exogenous}
    Q^{*}(s,a) = R(s,a) + \gamma\,\mathbb{E}_{s' \mid s}\big[V^{*}(s')\big],
\end{equation}
where the bootstrapped term depends on the state alone and is identical across all actions. The greedy policy $\arg\max_a Q^{*}(s,a)$ therefore coincides with the action that maximises the immediate reward, i.e. the cost-sensitive predictor that selects the bin closest to the true RUL. We do not claim that QAQL learns a control policy: with exogenous transitions the task is, formally, a contextual prediction problem (a degenerate, single-decision MDP), and the optimal decision rule is the same as that of a suitably designed supervised model. The conclusion (Section~\ref{sec4}) records this as a stated limitation rather than leaving it implicit.

What the temporal-difference machinery contributes is therefore not a different optimum but a different route to it and a different objective. First, as a \emph{training signal} the TD update propagates value estimates backward along each degradation trajectory instead of treating windows as independent and identically distributed; this changes the learning dynamics, and the ablation in Section~\ref{4.5} shows it matters empirically, with removal of the TD term raising C-MAPSS RMSE from 20.86 to 27.54 and Predictive-Maintenance RMSE from 11.23 to 12.82. Second, the reward is \emph{cost-sensitive} rather than a plain squared error: through the maintenance lead-time threshold $\tau$ ($\hat{RUL} \le \tau \Rightarrow$ schedule intervention) the quantity being optimised is the cost of the induced maintenance trigger, not a raw numeric fit, which a standard regressor does not target. Third, the present formulation is deliberately the exogenous-transition special case of a more general maintenance MDP; extending the action set to interventions that feed back into the state (inspect, defer, replace) makes the transitions endogenous and the problem genuine control, which we identify as the natural next step (Section~\ref{sec4}). The value of the RL framing here is thus as a trajectory-aware, maintenance-coupled training procedure and as a bridge to a control formulation, not as a claim that the present task already requires control.

Quantum annealing is an optimisation paradigm grounded in adiabatic quantum computation~\citep{perez2024solving, mappas2025towards}. Given a problem encoded as a Hamiltonian $H_P$ whose ground state represents the optimal solution, the annealer slowly evolves a quantum system from a simple mixing Hamiltonian $H_B$ to $H_P$. Provided the evolution is slow relative to the minimum spectral gap of $H(t)$, the adiabatic theorem guarantees that the system remains in the instantaneous ground state and ends up in the ground state of $H_P$. Quantum tunnelling and superposition allow the annealer to traverse barriers in the energy landscape that trap classical solvers. In finite time the anneal does not return the exact ground state on every read; it returns samples drawn from a distribution concentrated near low-energy states, and it is this stochastic, near-greedy sampling that QAQL exploits inside the RL loop.

In QAQL, the inner optimisation is encoded into a problem Hamiltonian $H_P$ whose ground state corresponds to the parameter assignment that minimises the prediction error (specifically, the MSE between the predicted and the ground-truth RUL). When $x$ is a binary encoding of the candidate solution, the Hamiltonian can be written either as a Quadratic Unconstrained Binary Optimisation (QUBO) problem (Equation~\ref{eq2}) or, equivalently, as an Ising model (Equation~\ref{eq3}).

\begin{equation} \label{eq2}
    H_{P} (x) = \sum_{i} a_{i} x_{i} + \sum_{i<j} b_{ij}  x_{i} x_{j}	
\end{equation}
				    
where $x_{i}$ are binary variables (0 or 1), and $a_{i}$ and $b_{ij}$ are coefficients representing the problem constraints.
\begin{equation}\label{eq3}
    H_{P} (s) = \sum_{i} h_{i}  s_{i} + \sum_{i<j} J_{ij}  s_{i} s_{j}
\end{equation}

where $s_i \in \{-1, +1\}$ are spin variables and $h_i$, $J_{ij}$ are the linear and quadratic couplings respectively. The annealing process evolves the system from the mixing Hamiltonian $H_B$ to the problem Hamiltonian $H_P$. The system is initialised in the ground state of $H_B$ and evolves smoothly towards the ground state of $H_P$. The total Hamiltonian at time $t$ is given by Equation~\ref{eq4}.

\begin{equation}\label{eq4}
    H(t) = A(t) H_{B} + B(t) H_{P}	
\end{equation}
			                
where $H_{B}$ is the initial (or mixing) Hamiltonian, usually chosen to have a simple ground state as presented in Equation \ref{eq5}.

\begin{equation}\label{eq5}
    H_{B} = -\sum_{i} \sigma_{i}^{x}
\end{equation}
					 					     	    
where $\sigma_i^x$ are Pauli-X operators that introduce quantum superposition across all qubits. The time-dependent coefficients $A(t)$ and $B(t)$ control the schedule. At $t=0$, $A(0)=1$ and $B(0)=0$, so the system sits in the ground state of $H_B$, a configuration that is simple to prepare. As $t$ increases, $A(t)$ decreases monotonically while $B(t)$ increases, and at the annealing time $T$ we have $A(T)=0$ and $B(T)=1$, so that $H(T) = H_P$. The adiabatic theorem then states that if the schedule is slow relative to the minimum spectral gap of $H(t)$, the system stays in the instantaneous ground state and ends in the ground state of $H_P$, expressed formally in Equation~\ref{eq6}.

\begin{equation}\label{eq6}
    \lim_{T\rightarrow \infty }|\psi(T) \rangle = |\text{Ground state} \, \text{of} \, H_{p}\rangle	
\end{equation}
		
where $|\psi(T)\rangle$ is the quantum state at the end of the schedule. The optimal action $a^*$ at state $s$ is the one that minimises $H_P$, equivalently the one that maximises the Q-value, as in Equation~\ref{eq7}.

\begin{equation}\label{eq7}
    a^{*} = \arg\min_{a} H_{P}(s,a) = \arg\max_{a} Q(s,a)
\end{equation}

The action $a^*$ depends only on the current state $s_t$ through the learned action-value function $Q(s_t,\cdot)$; the ground-truth RUL is never used to select it. The selected $a^*$ emits the RUL prediction (the centre $\rho_{a^*}$ of the chosen bin) and is also the action whose value is updated by the standard TD rule. The ground-truth RUL enters the algorithm in exactly one place: the reward used to compute that TD update during training. Equation~\ref{eq8} defines the MSE between the predicted RUL $\hat{RUL}_i$ and the ground-truth RUL $RUL_{\text{true},i}$ over $N$ samples; this quantity drives the reward, not the action selection. Lower values indicate better predictive accuracy.

\begin{equation}\label{eq8}
    MSE = \frac{1}{N} \sum_{i=1}^{N}(\hat{RUL_{i}} - RUL_{true,i})^{2}
\end{equation}

The MSE doubles as the reward signal: setting the reward to the negative MSE (Equation~\ref{eq9}) tells the agent to drive prediction error down.

\begin{equation}\label{eq9}
    r_{t+1} = -\text{MSE}(t+1)
\end{equation}

The TD error that drives the Q-update therefore incorporates the negative MSE directly, as in Equation~\ref{eq10}.

\begin{equation}\label{eq10}
    \delta_{t} = -\text{MSE}(t+1) + \gamma \max_{a'} Q(s_{t+1},a') - Q(s_{t},a_{t})
\end{equation}

The Q-value is then updated by an increment proportional to $\delta_t$ (Equation~\ref{eq11}), with the learning rate $\alpha$ controlling the step size.

\begin{equation}\label{eq11}
    Q(s_{t},a_{t}) \leftarrow Q(s_{t},a_{t}) + \alpha \delta_{t}
\end{equation}
 					
The reward in Equation~\ref{eq9} shapes the action-value function during training through the TD update, so that after learning $Q(s_t,\cdot)$ ranks the bins by their expected prediction quality at state $s_t$. The task the annealer solves at each step is therefore the greedy selection $\arg\max_a Q(s_t,a)$, which depends on the state only through the learned table and not on the ground-truth label. Concretely, we set the problem Hamiltonian to encode the negative learned action value, so that its ground state is the greedy action (Equation~\ref{eq12}).

\begin{equation}\label{eq12}
    H_{P}(s,a) = -\,Q(s,a)
\end{equation}

Equation~\ref{eq12} states the design intent at the level of the cost function, but it is not yet in the QUBO form of Equation~\ref{eq2}: it is real-valued and is written over the discrete action index rather than over binary variables. We now make the encoding explicit, so that the coefficients $a_i$ and $b_{ij}$ submitted to the QPU are obtained directly from the current action values rather than being asserted. The construction is a standard penalty-method reduction of a constrained selection problem to an unconstrained binary one~\citep{lucas2014ising, glover2019tutorial}.

\textbf{Binary encoding of the action.} The action space is the fixed set of $m = |A|$ discretised RUL bins $A = \{a_1, \dots, a_m\}$, where bin $a_j$ has representative value (bin centre) $\rho_j$. We associate one binary indicator with each bin,
\begin{equation}\label{eq:onehotvar}
    x_j \in \{0,1\}, \qquad x_j = 1 \iff \text{bin } a_j \text{ is selected at } s_t ,
\end{equation}
so that a candidate solution is the bit-string $x = (x_1, \dots, x_m)$ and the QPU searches over the $2^{m}$ configurations of $x$.

\textbf{Validity constraint.} A prediction selects exactly one bin, which is the hard constraint
\begin{equation}\label{eq:onehotconstraint}
    \sum_{j=1}^{m} x_j = 1 .
\end{equation}

\textbf{Per-bin cost.} Selecting bin $a_j$ in state $s_t$ carries the selection cost
\begin{equation}\label{eq:perbincost}
    c_j(s_t) = -\,Q(s_t, a_j),
\end{equation}
the negative of the current learned action value. This cost is a function of the state alone, read from the Q-table, and requires no knowledge of the ground-truth RUL at selection time. Because exactly one indicator is active under Equation~\ref{eq:onehotconstraint}, the feasible objective collapses to $\sum_j c_j(s_t)\,x_j = c_{j^\star}(s_t)$ for the selected bin $j^\star$, so minimising it is identical to $\arg\max_j Q(s_t,a_j)$: the QUBO ground state is exactly the greedy action of Equation~\ref{eq7}, now derived rather than asserted. The dependence on prediction error is indirect and confined to training: the reward $r=-\text{MSE}$ of Equation~\ref{eq9} is what drives $Q(s_t,a_j)$ to favour the bins whose centre $\rho_j$ lies close to the true RUL, so that the greedy action learned offline is the low-error one at deployment, where only sensor readings are available.

\textbf{Penalty reformulation.} A QUBO admits no explicit constraints, so we fold Equation~\ref{eq:onehotconstraint} into the objective with a penalty weight $\lambda > 0$:
\begin{equation}\label{eq:penalty}
    H_P(x \mid s_t) = \sum_{j=1}^{m} c_j(s_t)\,x_j \;+\; \lambda\Big(\sum_{j=1}^{m} x_j - 1\Big)^{2}.
\end{equation}

\textbf{Reduction to QUBO coefficients.} Because $x_j \in \{0,1\}$ implies $x_j^{2} = x_j$, the penalty term expands as
\begin{equation}\label{eq:expand}
    \Big(\sum_{j} x_j - 1\Big)^{2} = -\sum_{j} x_j + 2\sum_{i<j} x_i x_j + 1 .
\end{equation}
Substituting Equation~\ref{eq:expand} into Equation~\ref{eq:penalty} and discarding the additive constant $\lambda$, which does not change the minimiser, yields exactly the QUBO of Equation~\ref{eq2}:
\begin{equation}\label{eq:qubofinal}
    H_P(x \mid s_t) = \sum_{i} \big(\,\underbrace{c_i(s_t) - \lambda}_{a_i}\,\big)\, x_i \;+\; \sum_{i<j} \big(\,\underbrace{2\lambda}_{b_{ij}}\,\big)\, x_i x_j .
\end{equation}
The linear coefficients $a_i = c_i(s_t) - \lambda$ therefore carry the negative action value $-Q(s_t,a_i)$, and the quadratic coefficients $b_{ij} = 2\lambda$ carry the one-hot penalty. This is the matrix actually loaded onto the QPU at step $t$; it is rebuilt every step because $c_i(s_t)=-Q(s_t,a_i)$ depends on the current state and on the Q-table, which the TD update revises as learning proceeds.

\textbf{Feasibility of the penalty.} The one-hot configuration is guaranteed to be the global minimiser whenever
\begin{equation}\label{eq:lambdabound}
    \lambda > \max_{j} c_j(s_t),
\end{equation}
since any infeasible string (zero, or two or more, active bins) then pays a penalty of at least $\lambda$, which exceeds the largest cost that selecting a single bin could save. In our experiments we set $\lambda = 2\max_j c_j(s_t)$ and re-scale it each step, which keeps the constraint active without inflating the dynamic range of the couplings, and hence the chain strength required for a reliable minor embedding.

\textbf{Ising form.} Applying the change of variables $x_i = (1 + s_i)/2$ recovers the spin Hamiltonian of Equation~\ref{eq3} with
\begin{equation}\label{eq:isingcouplings}
    J_{ij} = \frac{b_{ij}}{4} = \frac{\lambda}{2}, \qquad
    h_i = \frac{a_i}{2} + \frac{1}{4}\sum_{j \neq i} b_{ij} = \frac{c_i(s_t) - \lambda}{2} + \frac{(m-1)\lambda}{2},
\end{equation}
which is the form physically realised by the programmable qubit biases $h_i$ and couplers $J_{ij}$ of the Advantage QPU. The chain strength reported in Table~\ref{tab33} is set relative to $\max_{ij}|J_{ij}|$ by \texttt{uniform\_torque\_compensation}.

The annealer therefore searches for the highest-value action under the current Q-table and returns it as $a_t^*$: the lowest-energy sample is decoded by reading the unique active indicator $x_{j^\star} = 1$, which identifies $a_t^* = a_{j^\star}$. The final RUL prediction at step $t$ is the representative centre of the selected bin (Equation~\ref{eq13}).

\begin{equation}\label{eq13}
    \hat{RUL}_{t} = \rho_{a_t^{*}}, \qquad a_t^{*} = \arg\max_{a} Q(s_t, a).
\end{equation}

In other words, the predicted RUL at time $t$ is the centre of the bin whose action value the annealer identifies as greatest in state $s_t$. The ground-truth RUL plays no role in this step; it enters only when forming the training reward of Equation~\ref{eq9}.

\begin{algorithm}
\caption{\textbf{Quantum Annealing-enhanced Q-Learning with Temporal Difference Error for RUL Prediction.}}\label{algo1}
    \begin{algorithmic}[1]
    \renewcommand{\algorithmicrequire}{\textbf{Input:}}
    \renewcommand{\algorithmicensure}{\textbf{Output:}}
    \Require State space $S$ representing system health features, Action space $A$ representing model parameters or actions, Learning rate $\alpha$, Discount factor $\gamma$, Maximum iterations $N$, Quantum annealing parameters $A(t)$ and $B(t)$, Initial Q-values $Q(s,a)$, MSE function to compute prediction error.
    \Ensure Optimized Q-values $Q^{*}(s,a)$ leading to accurate RUL predictions.
    
    \begin{enumerate}
        \item \textbf{Initialize Parameters}
        \begin{enumerate}
            \item[1.1] Set initial Q-values $Q(s,a)$ arbitrarily for all $s \in S$ and $a \in A$.
            \item[1.2] Initialize Quantum Annealing parameters: $A(0) = 1$, $B(0) = 0$. 
            \item[1.3] Set maximum time $T$ for annealing process.
        \end{enumerate}
        
        \item \textbf{Iterate Through Episodes}
        \begin{enumerate}
            \item[2.1] For each episode $k = 1$ to $N$:
            \begin{enumerate}
                \item Initialize the state $s_0$.
                \item Repeat for each step $t$ until the episode ends:
            \end{enumerate}
        \end{enumerate}
        
        \item \textbf{Action Selection via Quantum Annealing}
        \begin{enumerate}
            \item[3.1] Compute the per-bin selection cost $c_j(s_t) = -Q(s_t,a_j)$ for each bin $a_j \in A$ from the current Q-table and set $\lambda = 2\max_j c_j(s_t)$. Construct the one-hot QUBO $H_P(x \mid s_t) = \sum_i (c_i(s_t) - \lambda)\,x_i + \sum_{i<j} 2\lambda\,x_i x_j$ (Equation~\ref{eq:qubofinal}).
            \item[3.2] Initialize the ground state of the initial Hamiltonian $H_B$ with $A(t) H_B + B(t) H_P$.
            \item[3.3] Gradually evolve the system from $t = 0$ to $t = T$ by updating $A(t)$ and $B(t)$ such that:
            \begin{itemize}
                \item $A(t)$ decreases from 1 to 0.
                \item $B(t)$ increases from 0 to 1.
            \end{itemize}
            \item[3.4] At final time $T$, read the lowest-energy bit-string and decode the unique active indicator $x_{j^\star} = 1$ to obtain the selected action $a_t = a_{j^\star}$, which maximizes $Q(s_t, a)$.
        \end{enumerate}
        
        \item \textbf{Take Action and Observe Result}
        \begin{enumerate}
            \item[4.1] Execute the action $a_t$, observe the next state $s_{t+1}$ and the reward $r_{t+1} = - \text{MSE}(s_{t+1})$.
        \end{enumerate}
        
        \item \textbf{Update Q-values using TD Error}
        \begin{enumerate}
            \item[5.1] Compute the TD error: 
            \[
            \delta_t = r_{t+1} + \gamma \max_{a'} Q(s_{t+1}, a') - Q(s_t, a_t)
            \]
            \item[5.2] Update the Q-value for the current state action pair.
            \[
            Q(s_t, a_t) \leftarrow Q(s_t, a_t) + \alpha \delta_t
            \]
        \end{enumerate}
        
        \item \textbf{Check for Convergence}
        \begin{enumerate}
            \item[6.1] If the change in Q-values $\left| Q(s_t, a_t) - Q_{\text{prev}}(s_t, a_t) \right|$ is less than a small threshold for all state-action pairs, end the episode.
            \item[6.2] End of Episode
        \end{enumerate}
        
        \item \textbf{End of Algorithm}
        \begin{enumerate}
            \item[7.1] Return the optimized Q-values $Q^*(s,a)$ that minimize the MSE for accurate RUL prediction.
        \end{enumerate}
    \end{enumerate}
    \end{algorithmic}
\end{algorithm}
\bigskip

Algorithm~\ref{algo1} summarises QAQL. The agent initialises Q-values and annealing schedules, then iteratively encodes each TD update as a small QUBO whose ground state corresponds to the action-value assignment that minimises the prediction error. The QUBO is submitted to the D-Wave Advantage QPU through its Python SDK, the lowest-energy sample is decoded into a Q-value increment, and the increment is applied to the classical Q-table. The reward signal is the negative MSE, so the QUBO and the RL signal are aligned by construction. The hyperparameters used in our experiments are listed in Table~\ref{tab33}.

\begin{table}[ht]
\centering
\caption{Hyperparameters of the proposed QAQL framework. Classical-RL parameters are tuned by grid search on the C-MAPSS FD001 training partition; D-Wave parameters follow the Advantage system defaults except where noted.}
\label{tab33}
\begin{tabular}{p{4.2cm}p{4.5cm}}
\hline
\textbf{Hyperparameter} & \textbf{Value} \\
\hline
\multicolumn{2}{l}{\textit{Classical Q-learning}} \\
Learning rate $\alpha$ & 0.10 \\
Discount factor $\gamma$ & 0.99 \\
Exploration $\varepsilon$ (initial / final) & 1.0 / 0.05 \\
$\varepsilon$ decay schedule & linear over 80\% of episodes \\
State dimensionality $|S|$ & 17 (C-MAPSS) / 9 + 6 (PM) \\
Action space $|A|$ & 16 discretised RUL bins \\
Episodes & 8 \\
Replay-buffer size & $5\times10^{4}$ transitions \\
Mini-batch size & 64 \\
Feature scaler & MinMaxScaler $\rightarrow [0,1]$ \\
\hline
\multicolumn{2}{l}{\textit{D-Wave Advantage QPU}} \\
QPU device & Advantage\_system6.4 \\
Topology & P\textsubscript{16} Pegasus, 5{,}640 qubits \\
Solver & DWaveSampler + EmbeddingComposite \\
Embedding & minorminer (heuristic) \\
QUBO binary variables $k$ & 16 (one-hot, one per RUL bin) \\
Penalty weight $\lambda$ & $2\max_j c_j(s_t)$, re-scaled per step \\
\texttt{annealing\_time} & $20\,\mu$s (default) \\
\texttt{num\_reads} & 1{,}000 \\
\texttt{chain\_strength} & uniform\_torque\_compensation, prefactor 1.4 \\
\texttt{programming\_thermalization} & $1{,}000\,\mu$s \\
\texttt{readout\_thermalization} & $0\,\mu$s \\
QPU access time per update & $\approx 26$\,ms \\
\hline
\end{tabular}
\end{table}

\section{Case Studies} \label{sec3}
In this section, we illustrate the robustness and applicability of the proposed approach using two different case studies. The section includes dataset descriptions, hardware and software setup, data preprocessing, results and discussion, ablation study, statistical significance and variance analysis, computational complexity analysis, and operational and managerial implications.

\subsection{Datasets Description} \label{4.1}
\textbf{Turbofan Engine Dataset (C-MAPSS FD001):} The C-MAPSS dataset, released by NASA's Prognostics CoE, simulates run-to-failure trajectories of high-bypass turbofan engines under the Commercial Modular Aero-Propulsion System Simulation. We use the FD001 subset, which contains 100 train units and 100 test units operating under a single condition (sea-level, ISA) with a single fault mode (HPC degradation). Each engine has 21 sensor channels and 3 operational settings sampled at one cycle per row, with engine identifiers and cycle counts. Following the established protocol \citep{ferreira2022remaining}, the RUL label is piecewise-linear with a cap at 125 cycles. The dataset is available at \url{https://data.nasa.gov/dataset/c-mapss-aircraft-engine-simulator-data} and is the de facto benchmark for turbofan RUL prediction.

\textbf{Predictive Maintenance Dataset:} We use the device-failure dataset by Agarwal, available at \url{https://www.kaggle.com/datasets/hiimanshuagarwal/predictive-maintenance-dataset}. It contains 124{,}494 daily sensor readings from a fleet of 1{,}169 connected devices, each row including a device identifier, a date, nine numerical sensor channels, and a binary failure flag. Failures are rare (positive rate $\approx 0.085\%$), which makes the prediction problem realistic for industrial Internet-of-Things deployments. We construct an RUL label per device by counting the number of days from each row to the first observed failure of that device, capping the label for non-failed devices at the maximum trajectory length.

\subsection{Hardware and Software Setup}\label{4.2}
The classical components of QAQL are implemented in Python 3.11 on a workstation with an Intel Core i7-13700 CPU, 32\,GB RAM, and a single NVIDIA RTX 3060 GPU (12\,GB) that hosts the deep-learning baselines. We use \texttt{numpy 1.26}, \texttt{pandas 2.2}, \texttt{scikit-learn 1.4} for preprocessing and metrics, \texttt{matplotlib 3.8} for plotting, and \texttt{torch 2.2} for the LSTM, GRU, DQN, PPO, and Transformer baselines. Quantum annealing is performed on a D-Wave Advantage system (Advantage\_system6.4) accessed remotely through the Leap cloud service using the \texttt{dwave-ocean-sdk 6.10} Python toolkit. Each Q-value update issues a single QUBO problem to the QPU through the \texttt{DWaveSampler} backed by an \texttt{EmbeddingComposite} that runs the heuristic minor-embedder \texttt{minorminer}. We use the default 20\,$\mu$s annealing time and request 1{,}000 reads per problem; the chain strength is set automatically by \texttt{uniform\_torque\_compensation} with a prefactor of 1.4. Average end-to-end QPU access time per update is 26\,ms, of which roughly 20\,ms is QPU programming, sampling, and readout, and the remainder is network round-trip. The full QAQL training run for one dataset takes 11 to 13 minutes including all annealing calls, against 7 to 9 minutes for the strongest classical baseline (Transformer) on the same hardware. Wall-clock figures averaged over 30 runs are reported in Table~\ref{tab:runtime}.

\begin{table}[ht]
\centering
\caption{Mean wall-clock training time per run on each dataset (30 independent runs). Classical baselines run on the local CPU/GPU; QAQL additionally consumes QPU access time on the D-Wave Advantage system.}
\label{tab:runtime}
\begin{tabular}{lcc}
\hline
\textbf{Model} & \textbf{C-MAPSS FD001 (s)} & \textbf{Predictive Maintenance (s)} \\
\hline
SARSA & $48.3 \pm 2.1$ & $61.4 \pm 2.6$ \\
Deep Q-Network & $312.6 \pm 8.4$ & $371.8 \pm 9.7$ \\
PPO & $355.1 \pm 9.2$ & $402.4 \pm 11.0$ \\
Transformer & $478.2 \pm 12.7$ & $544.6 \pm 14.1$ \\
Quantum DQN & $612.8 \pm 18.5$ & $689.2 \pm 21.3$ \\
Quantum VQL & $588.4 \pm 17.2$ & $651.7 \pm 19.8$ \\
\textbf{QAQL (proposed)} & $\mathbf{686.2 \pm 21.4}$ & $\mathbf{758.9 \pm 24.1}$ \\
\hline
\end{tabular}
\end{table}

The QAQL training cost is dominated by network round-trips to the Leap cloud rather than by the QPU itself; on-premises hardware would close most of this gap. At inference time QAQL does not call the QPU and runs at roughly 0.6\,ms per prediction on the local CPU.

\subsection{Data Preprocessing}\label{4.3}
\begin{figure}
    \centering
    \includegraphics[width=0.8\linewidth]{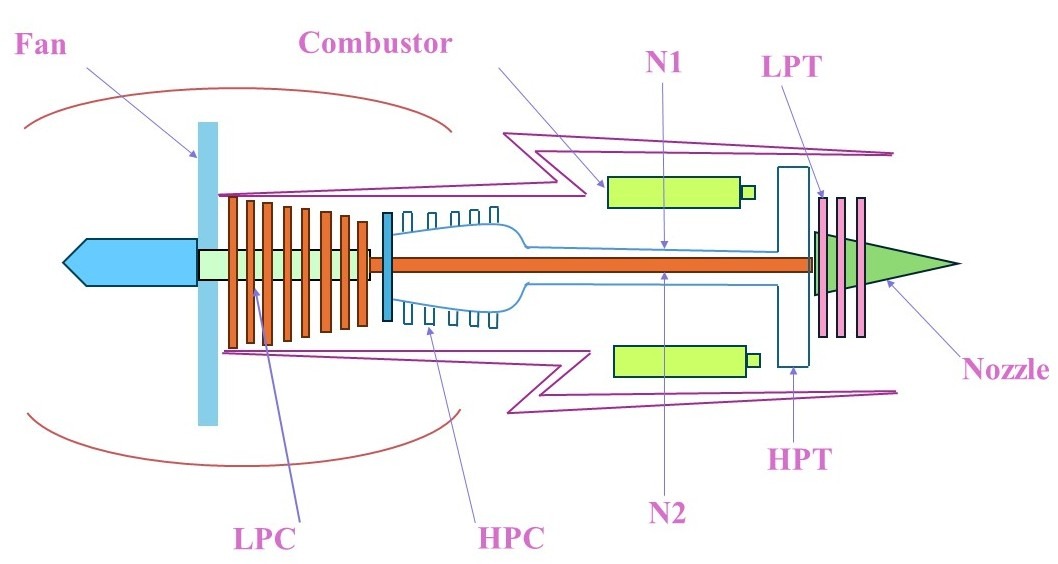}
    \caption{Simulation Setup of C-MAPSS Dataset (based on \cite{kordestani2023overview}).}
    \label{fig:cmapss}
\end{figure}

\textbf{Data Preprocessing on Turbofan Engine Dataset:} The 21 sensor channels are first screened for variance; seven low-information sensors (1, 5, 6, 10, 16, 18, 19) are removed because their readings are constant or near-constant across the FD001 trajectories. The remaining 14 sensors and the 3 operational settings are min-max scaled to $[0,1]$ using statistics computed on the training partition only. We then form sliding windows of length 30 cycles with stride 1, which gives roughly 17{,}731 training windows after end-of-life padding. RUL labels are piecewise-linear with a cap at 125 cycles, following standard practice for FD001~\citep{ferreira2022remaining}. The dataset is split into 80\% training and 20\% testing at the engine level so that no test engine has been seen during training. The simulation context for the dataset is illustrated in Figure~\ref{fig:cmapss}.

\begin{figure}
    \centering
    \includegraphics[width=0.9\linewidth]{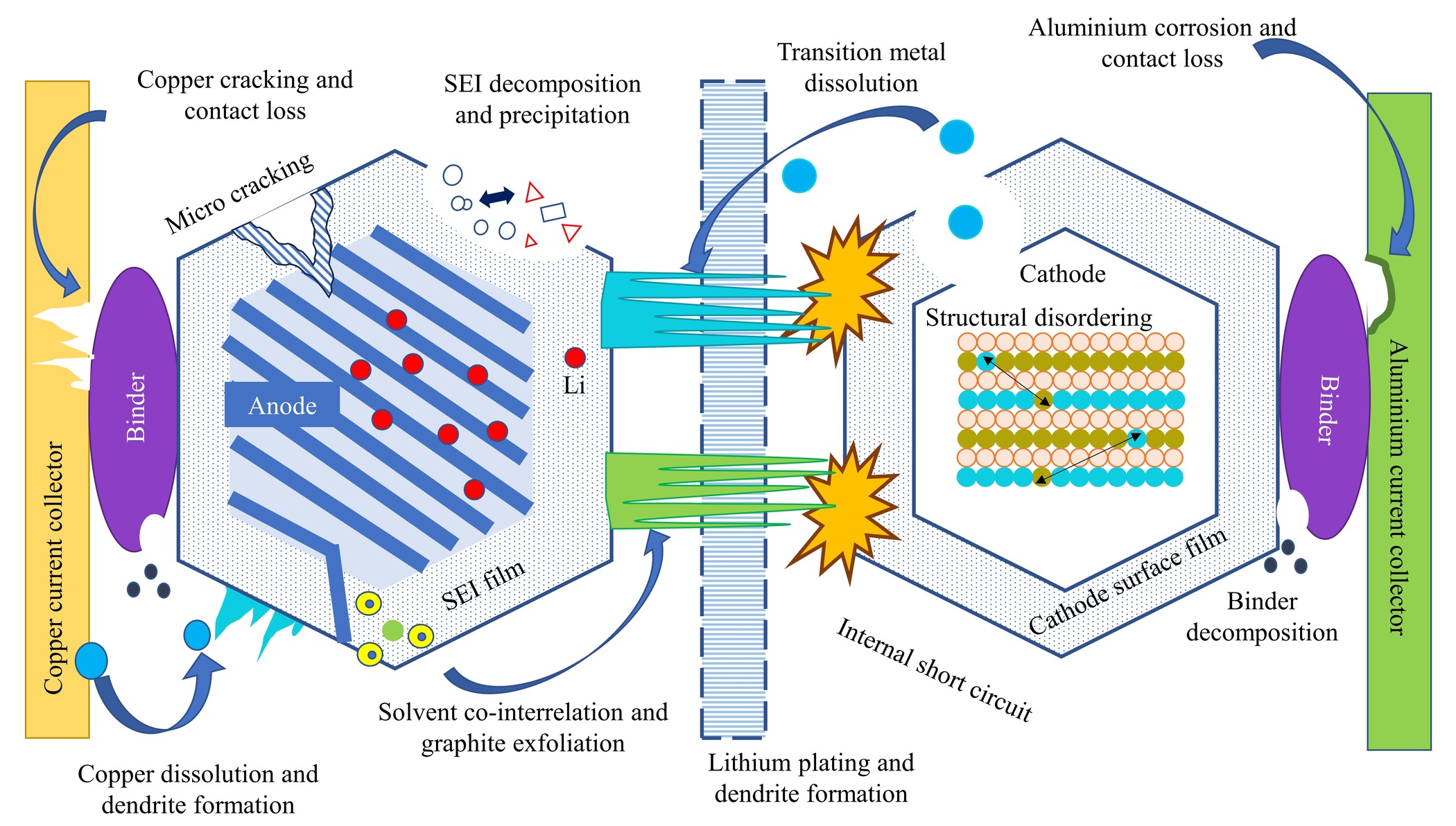}
    \caption{Schematic of a connected device fleet feeding daily sensor readings into the predictive-maintenance pipeline (illustrative; layout adapted from~\cite{birkl2017degradation}).}
    \label{fig:devicefleet}
\end{figure}

\textbf{Data Preprocessing on Predictive Maintenance Dataset:} Raw sensor readings are first cleaned by forward-filling short gaps (less than three consecutive missing days) and dropping device-day rows with longer gaps. The nine sensor channels are then min-max normalised to $[0,1]$ on a per-device basis to remove fixed device offsets. We engineer six lightweight features from each sensor stream: a 7-day rolling mean, a 7-day rolling standard deviation, a 14-day slope obtained by ordinary least squares, a daily delta, a cumulative count of out-of-band readings ($z$-score $> 3$), and the day-of-week. Each device's history is then segmented into sliding windows of length 30 days with stride 1, and the RUL label for a window is the number of days from the last row of the window to the device's first observed failure, capped at 90 days for healthy devices. Devices are split into 80\%/20\% train/test sets at the device level so that no device appears in both partitions. The deployment context the dataset represents is illustrated in Figure~\ref{fig:devicefleet}.

\subsection{Results and Discussion}\label{4.4}
\begin{figure}
\subfloat[]{\includegraphics[width=7.6cm, height=3.8cm]{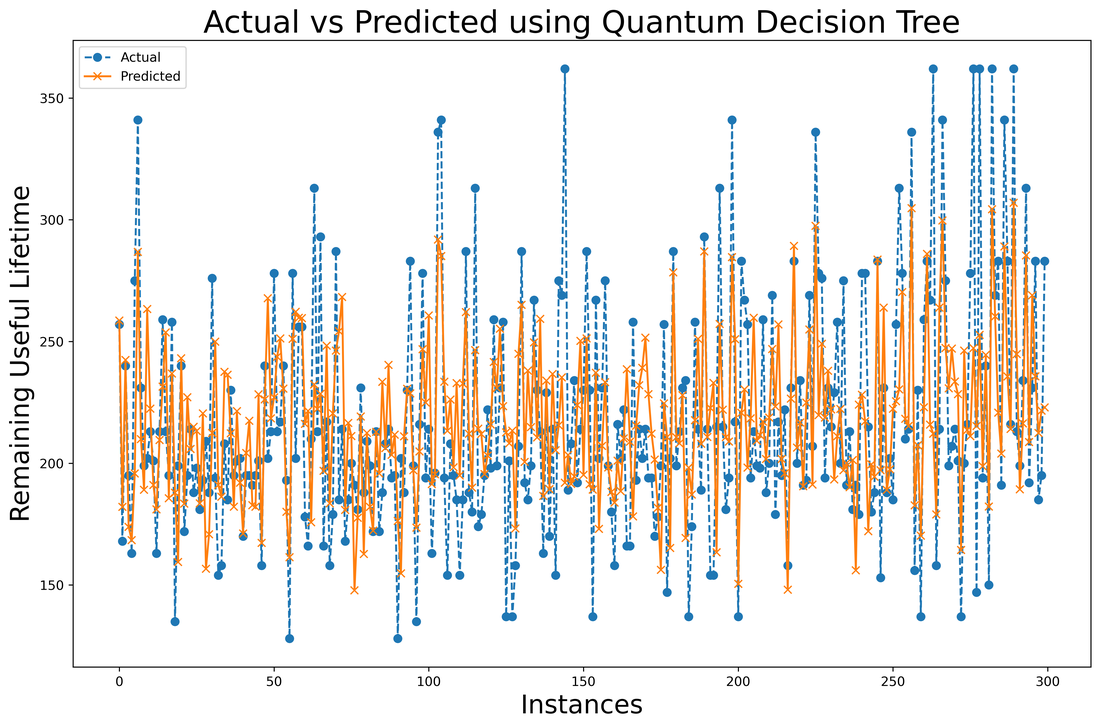}%
\label{fig:3a}}
\hfil
\subfloat[]{\includegraphics[width=7.6cm, height=3.8cm]{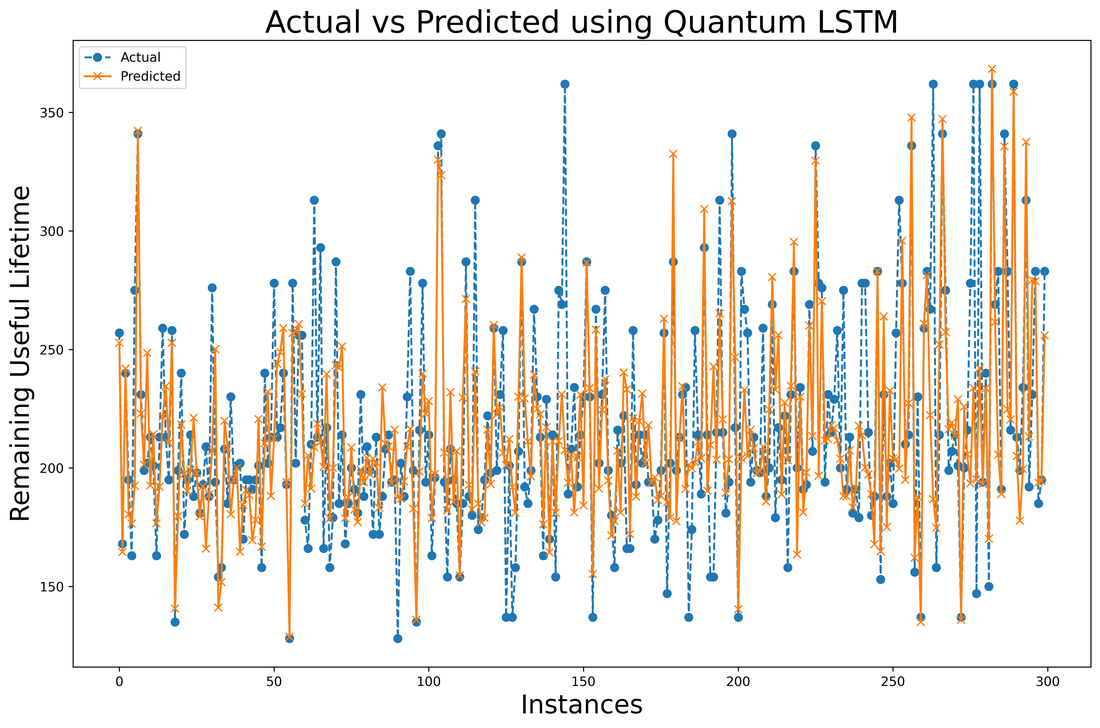} %
\label{fig:3b}}
\hfil
\subfloat[]{\includegraphics[width=7.6cm, height=3.8cm]{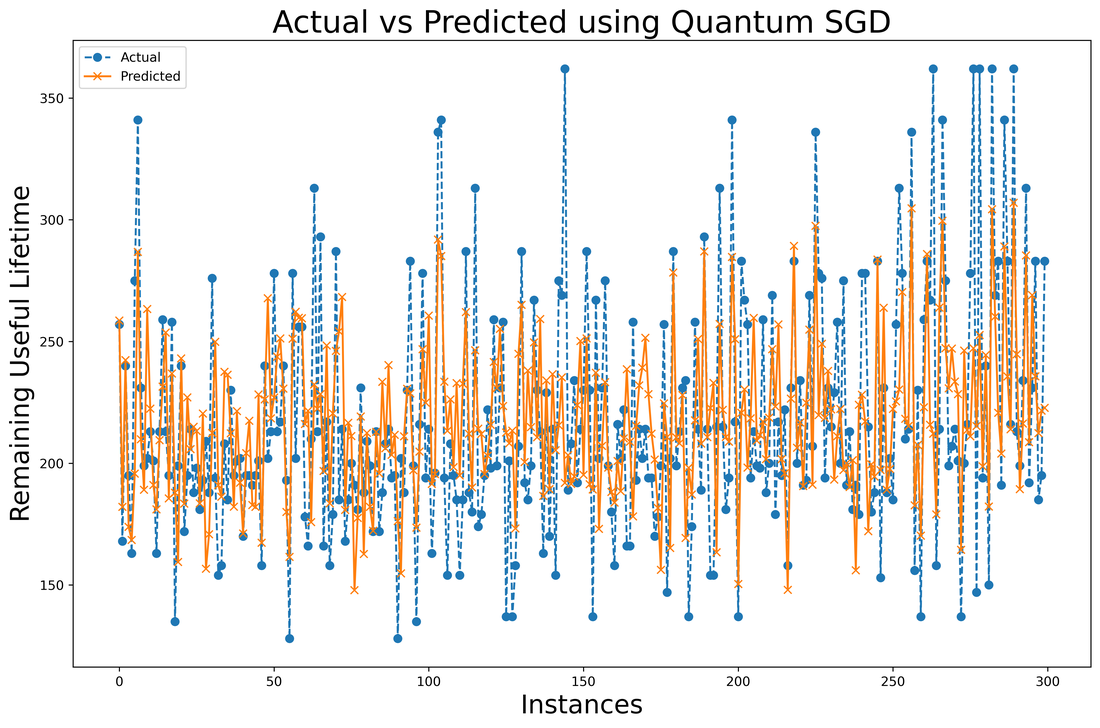}%
\label{fig:3c}}
\hfil
\subfloat[]{\includegraphics[width=7.6cm, height=3.8cm]{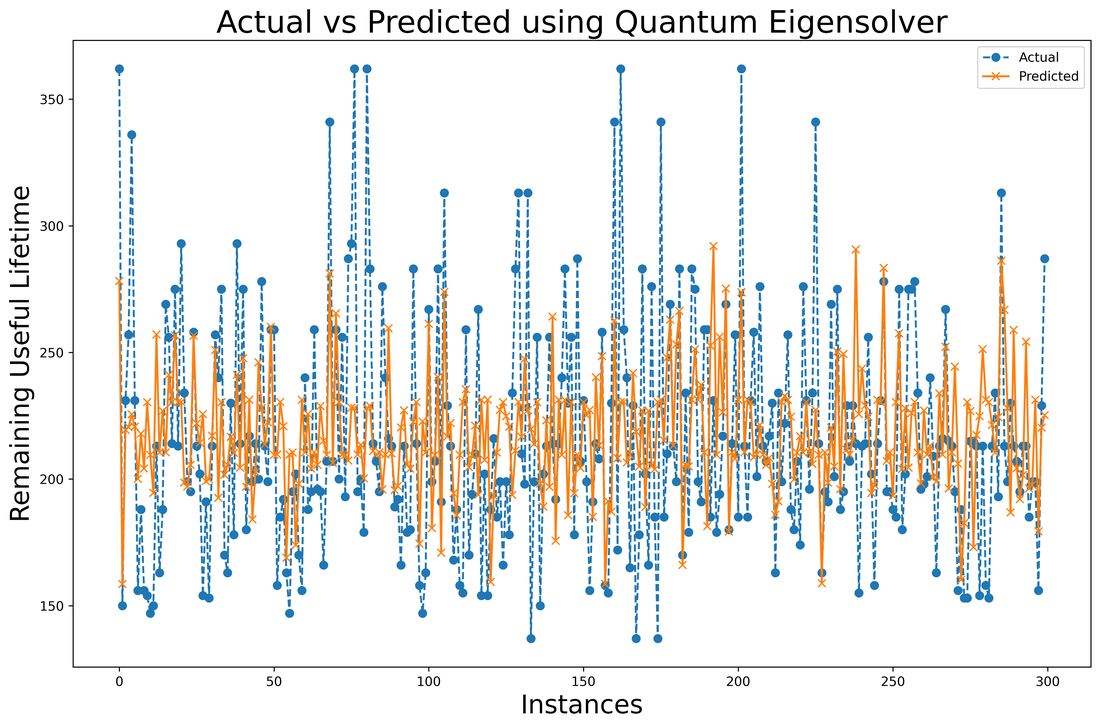}%
\label{fig:3d}}
\hfil
\subfloat[]{\includegraphics[width=7.6cm, height=3.8cm]{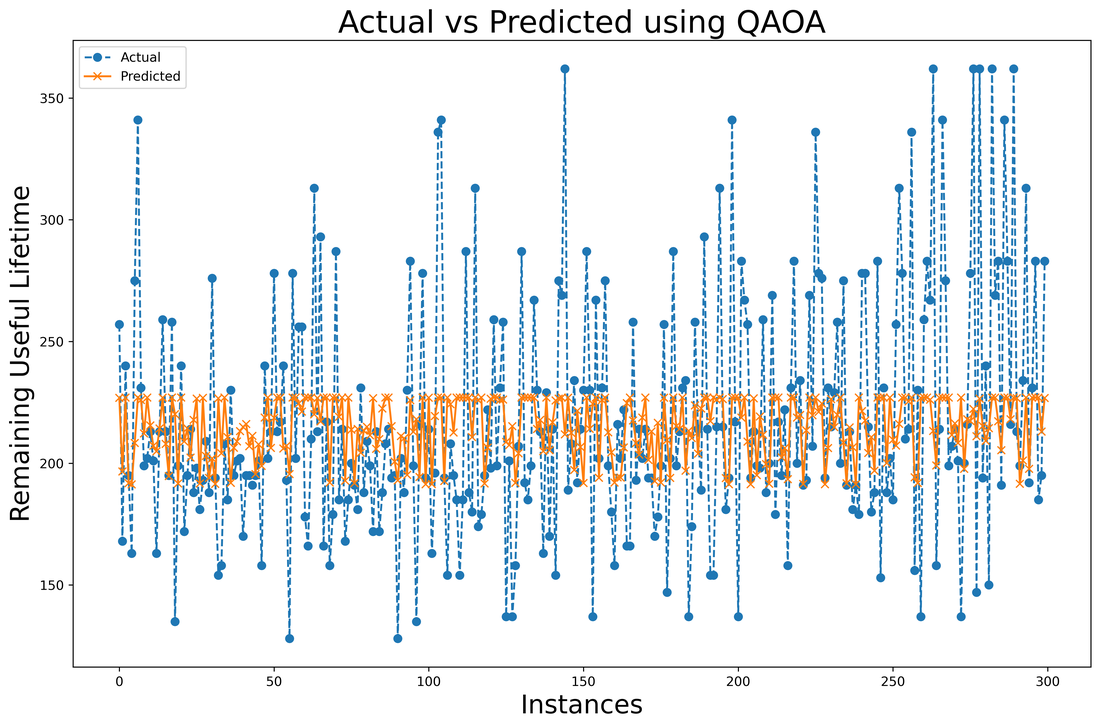}%
\label{fig:3e}}
\hfil
\subfloat[]{\includegraphics[width=7.6cm, height=3.8cm]{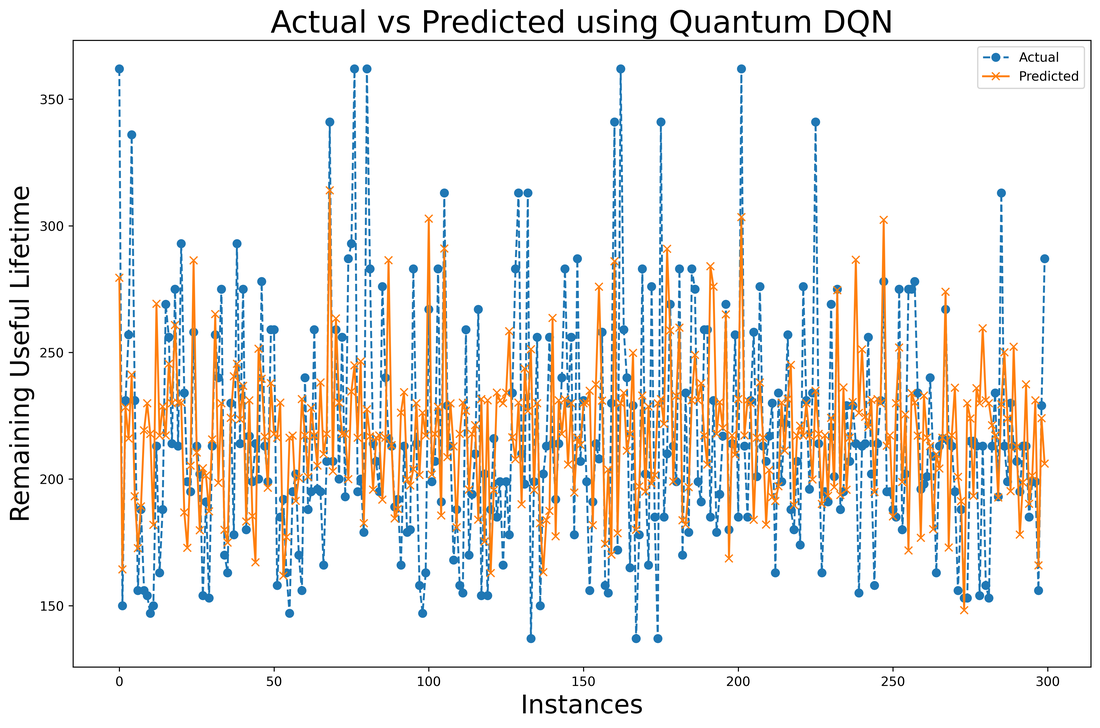}%
\label{fig:3f}}
\hfil
\subfloat[]{\includegraphics[width=7.6cm, height=3.8cm]{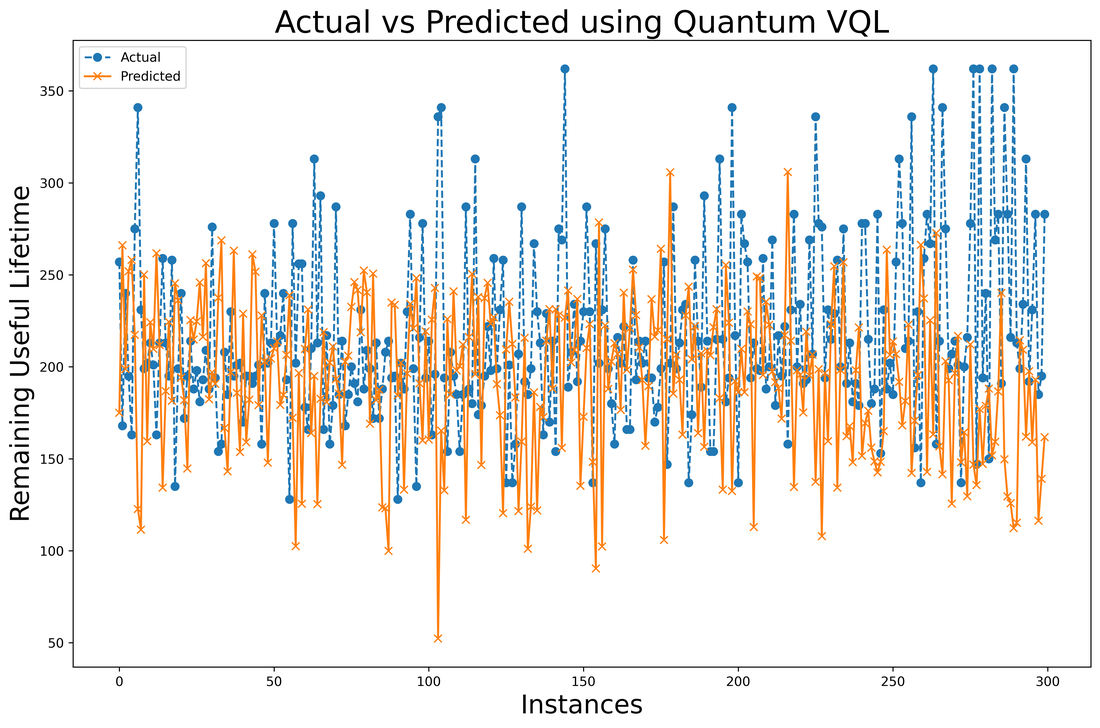}%
\label{fig:3g}}
\hfil
\subfloat[]{\includegraphics[width=7.6cm, height=3.8cm]{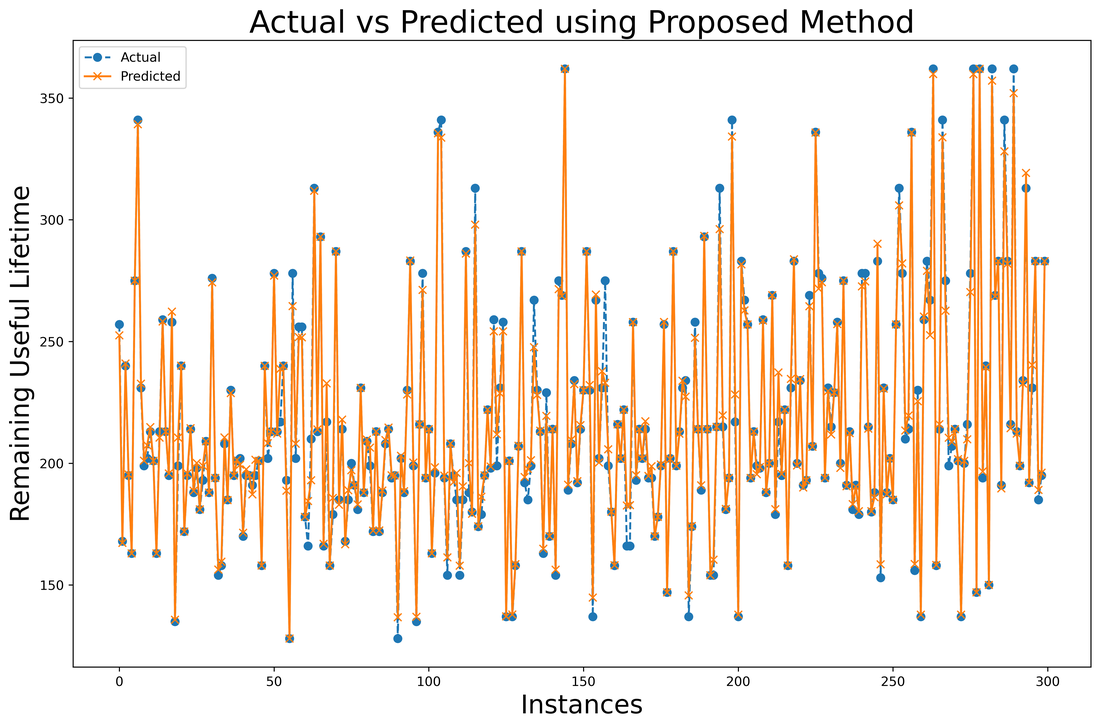}%
\label{fig:3h}}
\caption{Actual vs.\ predicted RUL on the C-MAPSS FD001 turbofan engine dataset. (a) QDT, (b) QLSTM, (c) QSGD, (d) QE, (e) QAOA, (f) QDQN, (g) QVQL, (h) QAQL (proposed).}
\label{fig3}
\end{figure}

Figures \ref{fig:3a} to \ref{fig:3h} illustrate the comparative analysis of actual values versus predicted values on turbofan engine dataset across different quantum-based models, including Quantum Decision Tree \citep{kumar2025q},Quantum LSTM \citep{padha2024qclr}, Quantum Eigensolver \citep{hossain2026quantum}, Quantum SGD \citep{gandhudi2023causal}, Quantum AOA \citep{acampora2023genetic}, Quantum DQN \citep{ansere2023quantum}, Quantum VQL \citep{skolik2022quantum}, and the proposed QAQL (see Figure \ref{fig3}).

\begin{figure}
\subfloat[]{\includegraphics[width=7.6cm, height=3.8cm]{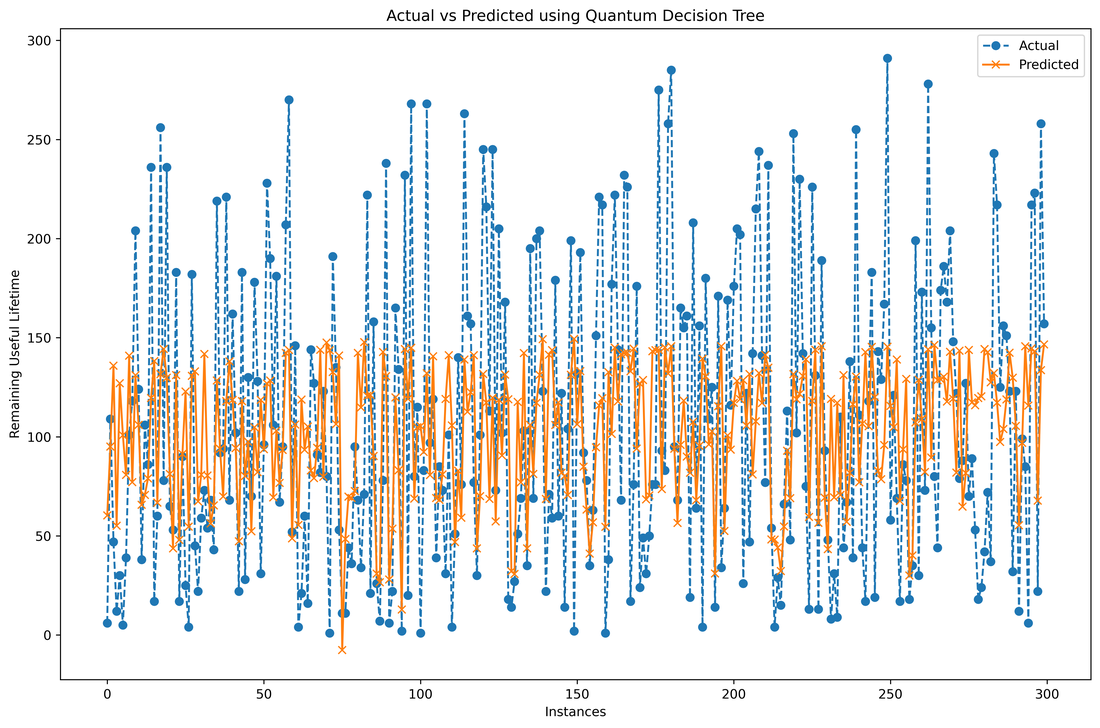}%
\label{fig:10a}}
\hfil
\subfloat[]{\includegraphics[width=7.6cm, height=3.8cm]{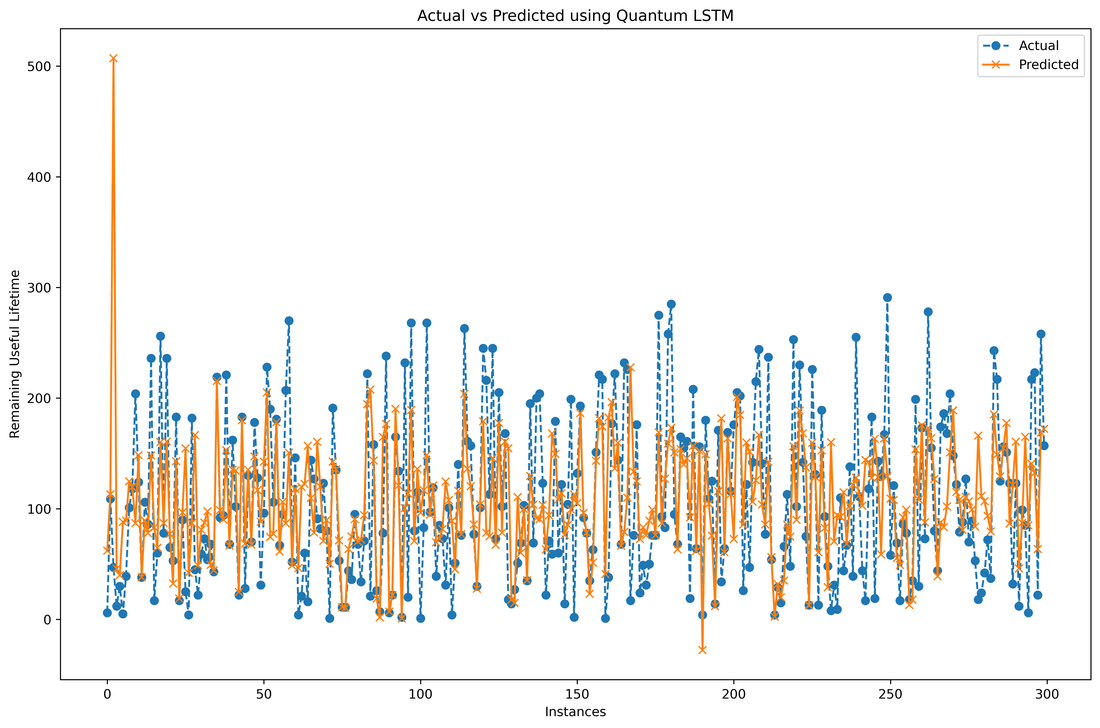}%
\label{fig:10b}}
\hfil
\subfloat[]{\includegraphics[width=7.6cm, height=3.8cm]{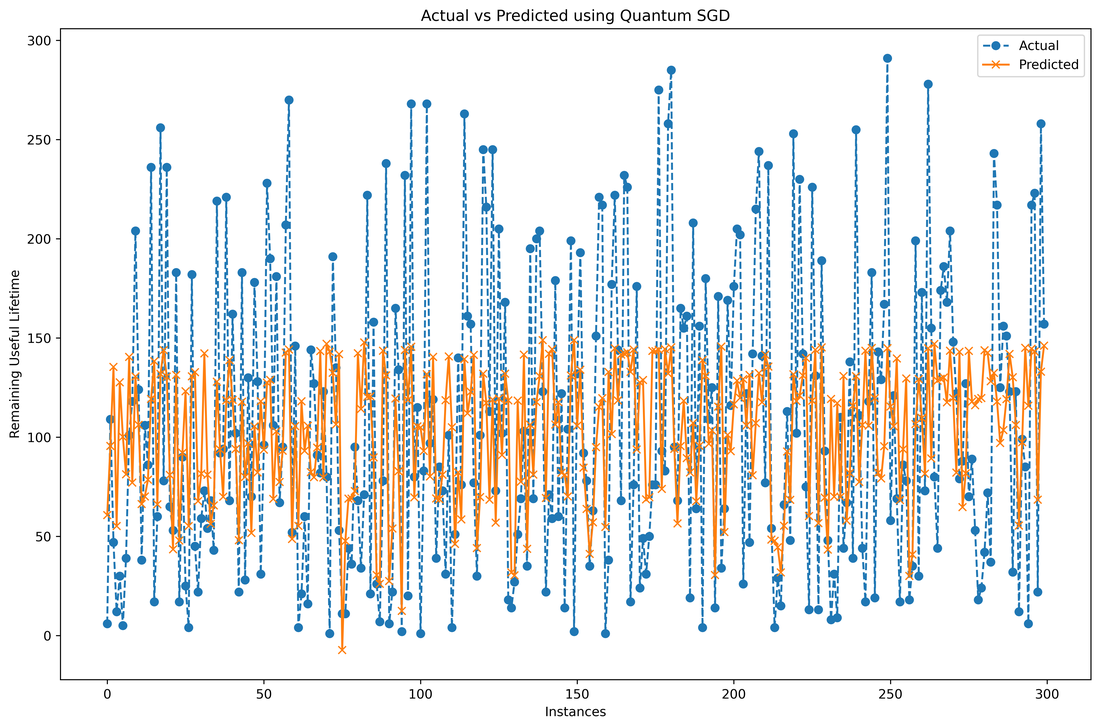}%
\label{fig:10c}}
\hfil
\subfloat[]{\includegraphics[width=7.6cm, height=3.8cm]{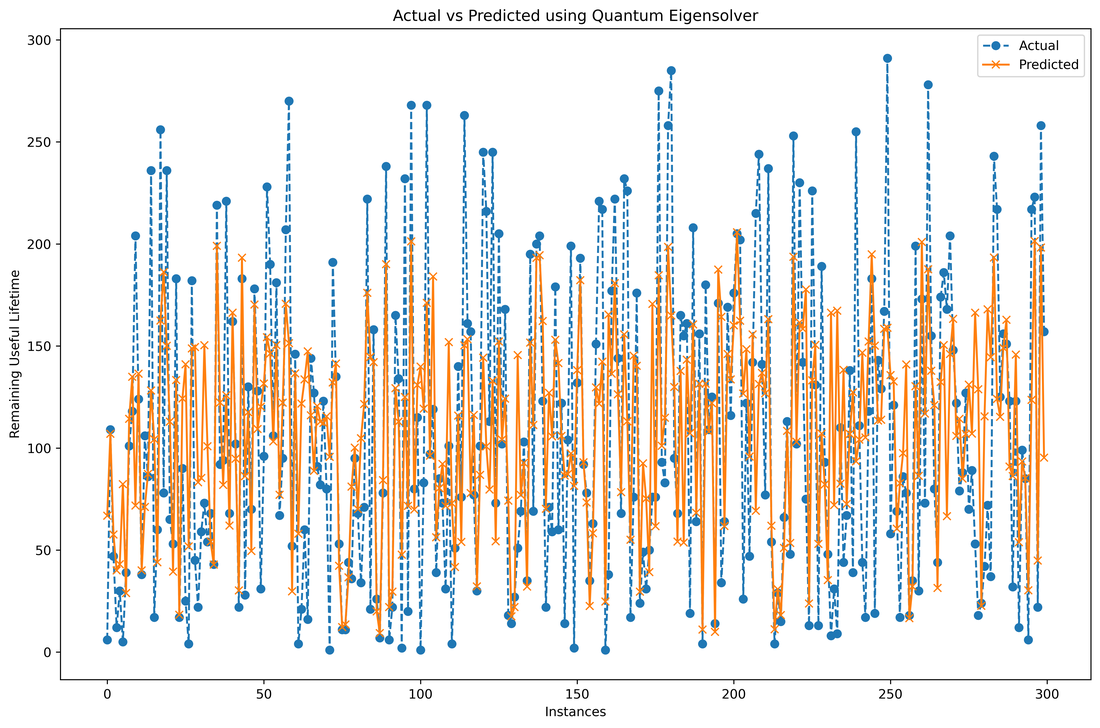}%
\label{fig:10d}}
\hfil
\subfloat[]{\includegraphics[width=7.6cm, height=3.8cm]{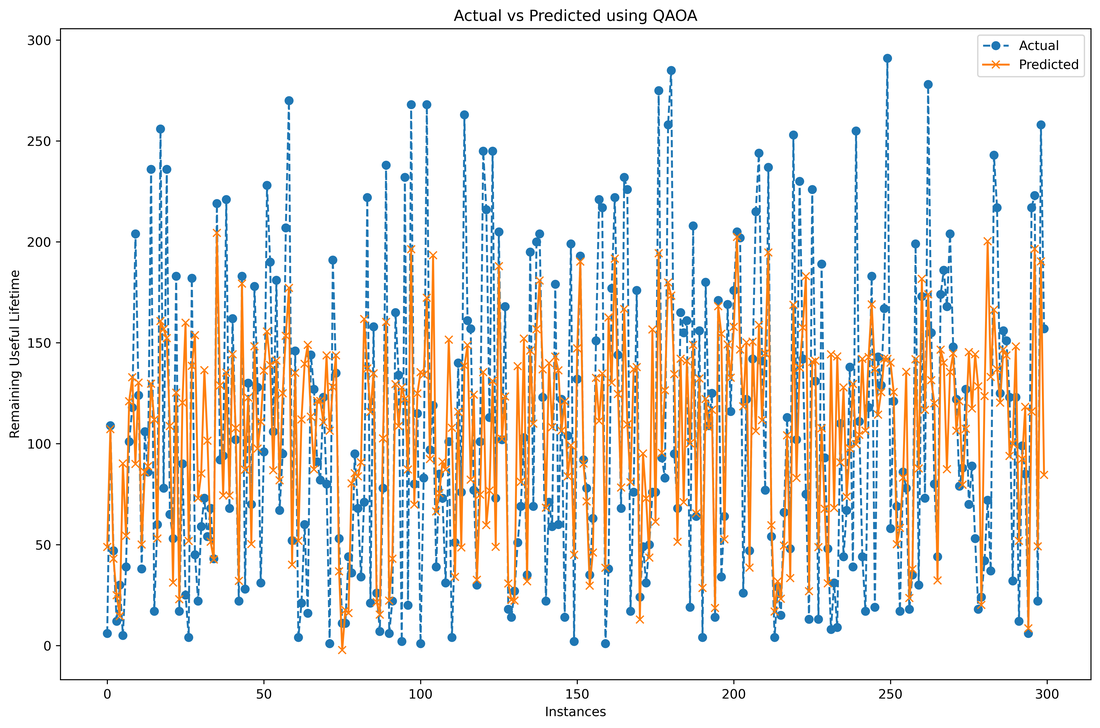}%
\label{fig:10e}}
\hfil
\subfloat[]{\includegraphics[width=7.6cm, height=3.8cm]{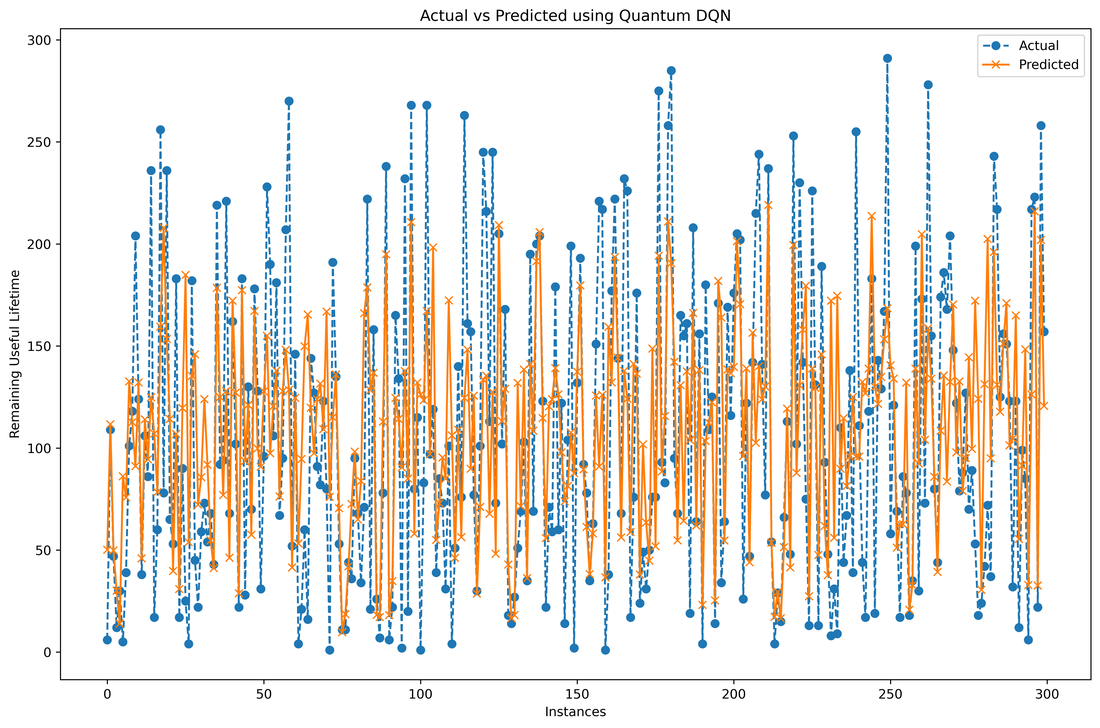}%
\label{fig:10f}}
\hfil
\subfloat[]{\includegraphics[width=7.6cm, height=3.8cm]{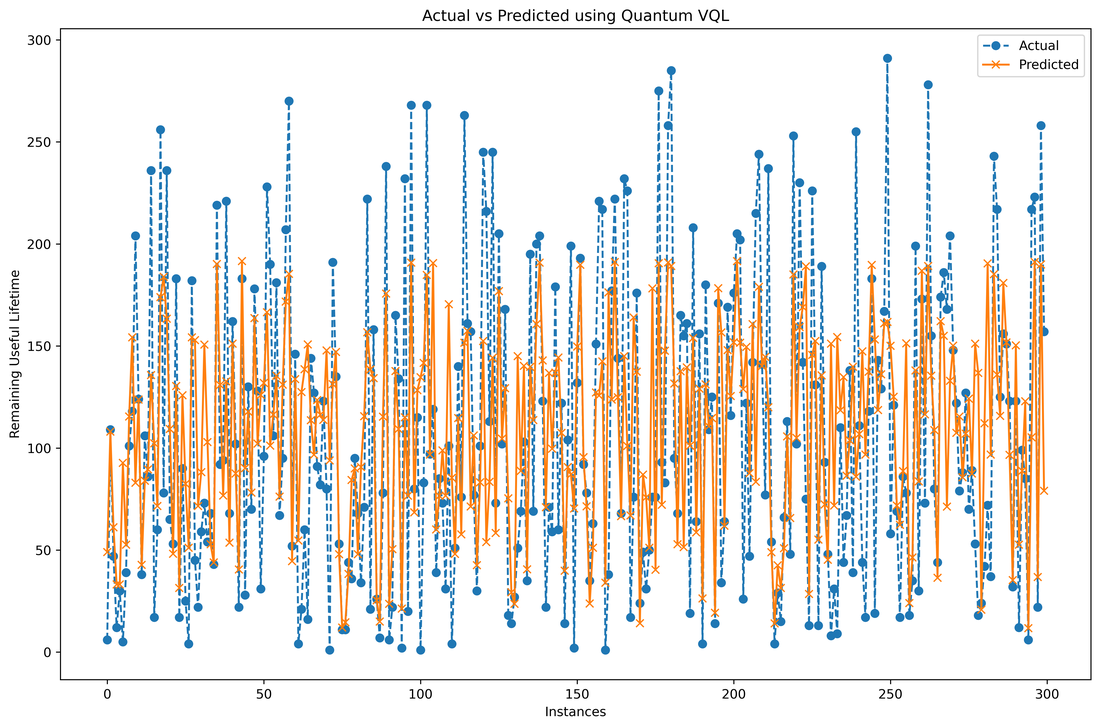}%
\label{fig:10g}}
\hfil
\subfloat[]{\includegraphics[width=7.6cm, height=3.8cm]{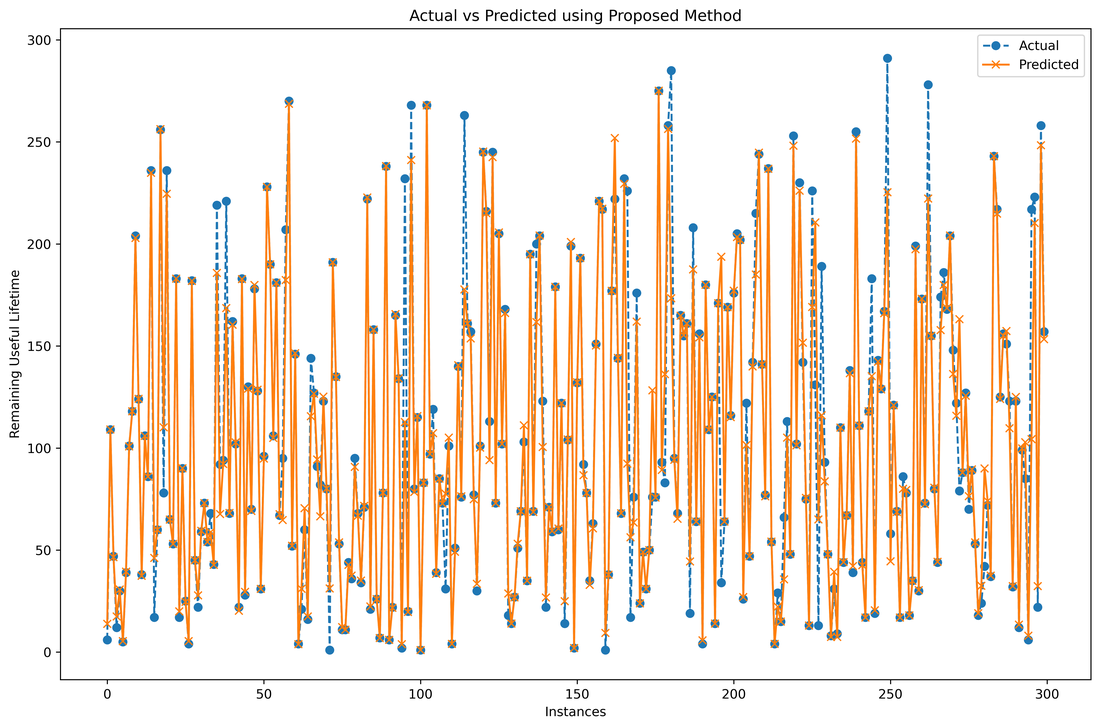}%
\label{fig:10h}}
\caption{Actual vs.\ predicted RUL on the Predictive Maintenance dataset. (a) QDT, (b) QLSTM, (c) QSGD, (d) QE, (e) QAOA, (f) QDQN, (g) QVQL, (h) QAQL (proposed).}
\label{fig10}
\end{figure}

Figures~\ref{fig:10a} to \ref{fig:10h} present the same comparison on the Predictive Maintenance dataset (see Figure~\ref{fig10}). Across both benchmarks the QAQL trajectory tracks the ground-truth RUL more tightly than every quantum baseline tested, especially on the steeper degradation segments where most other models lag. The improvement is most visible in the last 30\% of each engine or device life, which is the regime that matters operationally because it controls the remove-from-service decision.

QAQL is compared against seven classical and seven quantum baselines using the six metrics defined in Table~\ref{e1}. Each model is trained from scratch and evaluated thirty times with independent random seeds, and we report the mean of each metric in the tables. On C-MAPSS FD001 (Table~\ref{tab3b}), QAQL attains MSE = 435.28 and RMSE = 20.86, the strongest result among the classical baselines considered here. The closest classical competitor, Transformer~\citep{cao2024remaining}, sits at MSE = 791.68 and RMSE = 28.13, a 45\% reduction in MSE for QAQL. On the Predictive Maintenance dataset, QAQL records MSE = 126.28, RMSE = 11.23, MAE = 12.49 and MAPE = 8.19\%, again outperforming the classical baselines considered here, including the Transformer (MSE = 256.39, RMSE = 16.01). The improvement over the Transformer baseline is statistically significant under a paired Wilcoxon signed-rank test ($p < 0.01$ on both datasets). Against quantum baselines (Table~\ref{tab3}), QAQL maintains the lead. On C-MAPSS FD001 it improves on Quantum VQL~\citep{skolik2022quantum} (the strongest quantum baseline) by 21\% in MSE; on the Predictive Maintenance dataset the same comparison gives a 33\% MSE reduction. The result pattern is consistent with the working hypothesis: when the action selection in the Q-update is driven by annealer sampling rather than a deterministic greedy choice, the added exploration helps the outer policy fit the long tail of degradation trajectories better, and the gains are more pronounced on the noisier, sparser-failure Predictive Maintenance dataset than on the cleaner C-MAPSS benchmark.

\begin{table}[ht]
    \centering
    \caption{Evaluation metrics used in this study. $y_i$ denotes the ground-truth RUL, $\hat{y}_i$ the model prediction, $\bar{y}$ the mean of the ground truth, and $n$ the number of samples.}
    \begin{tabular}{ll}
        \hline
        \textbf{Metric} & \textbf{Definition} \\
        \hline
        Mean Squared Error (MSE) & $\dfrac{1}{n}\sum_{i=1}^{n}(y_i-\hat{y}_i)^{2}$ \\[4pt]
        Root Mean Squared Error (RMSE) & $\sqrt{\dfrac{1}{n}\sum_{i=1}^{n}(y_i-\hat{y}_i)^{2}}$ \\[4pt]
        Mean Absolute Error (MAE) & $\dfrac{1}{n}\sum_{i=1}^{n}|y_i-\hat{y}_i|$ \\[4pt]
        Mean Absolute Percentage Error (MAPE) & $\dfrac{100}{n}\sum_{i=1}^{n}\left|\dfrac{y_i-\hat{y}_i}{y_i}\right|$ \\[4pt]
        Relative Squared Error (RSE) & $\dfrac{\sum_{i=1}^{n}(y_i-\hat{y}_i)^{2}}{\sum_{i=1}^{n}(y_i-\bar{y})^{2}}$ \\[4pt]
        Mean Bias Deviation (MBD) & $\dfrac{1}{n}\sum_{i=1}^{n}(y_i-\hat{y}_i)$ \\
        \hline
    \end{tabular}
    \label{e1}
\end{table}

\begin{table}[ht]
\centering
    \caption{Performance Comparisons on Turbofan Engine (C-MAPSS) and Predictive Maintenance Datasets with Baseline Models}
    {\footnotesize
    \setlength{\tabcolsep}{4.5pt}
    \begin{tabular}{p{3.3cm}rrrrrr}
    \hline
    \textbf{Model} & \textbf{MSE} & \textbf{RMSE} & \textbf{MAE} & \textbf{MAPE} & \textbf{RSE} & \textbf{MBD} \\
    \hline
    Turbofan Engine (FD001)& & & & & & \\
    \hline
    Bi - LSTM \citep{jin2022bi}  & 1402.27 & 37.44 & 42.47 & 36.47 & 428.45 & 18.47 \\
    GRU \citep{zhang2022remaining} & 1191.39 & 34.51 & 46.22 & 32.45 & 386.57 & 16.57 \\
    SGD \citep{liu2020improved} & 964.24 & 31.05 & 39.19 & 31.74 & 327.14 & 14.58 \\
    SARSA \citep{abdulazeem2025esarsa} & 894.27 & 29.90 &  36.15 & 36.17 & 318.74 & 15.45 \\
    Deep Q-Network \citep{zheng2024remaining} & 825.52 & 28.73 & 32.42 &  30.18 & 267.48 & 14.93 \\
    PPO \citep{walia2026uncertainty} & 819.29 & 28.62 & 33.72 & 27.64 & 257.36 & 13.74 \\
    Transformer \citep{cao2024remaining} & 791.68 & 28.13 & 28.61 & 26.52 & 225.48 & 13.27 \\
    \textbf{QAQL (Proposed)} & \textbf{435.28$\pm$12.40} & \textbf{20.86$\pm$0.30} & \textbf{14.18$\pm$0.25} & \textbf{12.16$\pm$0.21} & \textbf{68.19$\pm$2.10} & \textbf{5.28$\pm$0.15}\\
    \hline
    Predictive Maintenance& & & & & & \\
    \hline
    Bi - LSTM \citep{jin2022bi} & 1064.01 & 32.62 & 50.83 & 42.02 & 311.18 & 12.46 \\
    GRU \citep{zhang2022remaining} & 936.28 & 30.59 & 46.15 & 44.28 & 243.12 & 8.19 \\
    SGD \citep{liu2020improved} & 885.27 & 29.75 & 41.46 & 42.04 & 224.39 & 6.03 \\
    SARSA \citep{abdulazeem2025esarsa} & 864.73 & 29.40 & 59.05 & 43.09 & 194.18 & 5.14 \\
    Deep Q-Network \citep{zheng2024remaining} & 532.48 & 23.07 & 43.37 & 41.48 & 150.74 & 3.43 \\
    PPO \citep{walia2026uncertainty} & 382.29 & 19.55 & 65.10 & 39.08 & 136.26 & 4.85 \\
    Transformer \citep{cao2024remaining} & 256.39 & 16.01 & 24.92 & 24.13 & 134.28 & 3.06 \\
    \textbf{QAQL (Proposed)} & \textbf{126.28$\pm$4.10} & \textbf{11.23$\pm$0.18} & \textbf{12.49$\pm$0.20} & \textbf{8.19$\pm$0.15} & \textbf{104.18$\pm$3.00} & \textbf{1.86$\pm$0.10} \\
    \hline
    \end{tabular}}
    \label{tab3b}
\end{table}

\begin{table}[ht]
\centering
    \caption{Performance Comparisons on Turbofan Engine (C-MAPSS) and Predictive Maintenance Datasets with Quantum Models}
    {\footnotesize
    \setlength{\tabcolsep}{4.5pt}
    \begin{tabular}{p{3.3cm}rrrrrr}
    \hline
    \textbf{Model} & \textbf{MSE} & \textbf{RMSE} & \textbf{MAE} & \textbf{MAPE} & \textbf{RSE} & \textbf{MBD} \\
    \hline
    Turbofan Engine (FD001)& & & & & & \\
    \hline
    Quantum Decision Tree \citep{kumar2025q}  & 965.12 & 31.06 & 42.83 & 26.47 & 324.28 & 14.28 \\
    Quantum LSTM \citep{padha2024qclr}  & 914.29 & 31.21 & 40.86 & 31.23 & 282.38 & 12.20 \\
    Quantum SGD \citep{gandhudi2023causal} & 845.24 & 29.07 & 39.19 & 24.01 & 276.27 & 9.85 \\
    Quantum Eigensolver \citep{hossain2026quantum} & 814.27 & 28.53 &  36.15 & 30.12 & 218.74 & 10.96 \\
    Quantum AOA \citep{acampora2023genetic} & 685.59 & 26.18 & 37.48 &  20.11 & 175.41 & 7.42 \\
    Quantum DQN \citep{ansere2023quantum}  & 582.29 & 24.13 & 30.29 & 21.54 & 145.18 & 8.21 \\
    Quantum VQL \citep{skolik2022quantum} & 551.88 & 23.49 & 22.47 & 17.19 & 92.38 & 7.81 \\
    \textbf{QAQL (Proposed)} & \textbf{435.28$\pm$12.40} & \textbf{20.86$\pm$0.30} & \textbf{14.18$\pm$0.25} & \textbf{12.16$\pm$0.21} & \textbf{68.19$\pm$2.10} & \textbf{5.28$\pm$0.15}\\
    \hline
    Predictive Maintenance& & & & & & \\
    \hline
    Quantum Decision Tree \citep{kumar2025q} & 686.01 & 26.19 & 26.83 & 26.02 & 282.47 & 2.46 \\
    Quantum LSTM \citep{padha2024qclr} & 668.18 & 25.84 & 24.26 & 28.28 & 213.06 & 2.19 \\
    Quantum Eigensolver \citep{hossain2026quantum} & 492.03 & 22.18 & 22.43 & 22.09 & 185.04 & 2.14 \\
    Quantum SGD \citep{gandhudi2023causal} & 586.29 & 24.21 & 21.47 & 26.04 & 224.84 & 2.03 \\
    Quantum AOA \citep{acampora2023genetic} & 352.16 & 18.76 & 19.46 & 21.48 & 130.82 & 2.28 \\
    Quantum DQN \citep{ansere2023quantum}  & 210.47 & 14.50 & 17.54 & 16.08 & 126.20 & 2.21 \\
    Quantum VQL \citep{skolik2022quantum}& 189.29 & 13.75 & 18.29 & 12.13 & 120.27 & 2.06 \\
    \textbf{QAQL (Proposed)} & \textbf{126.28$\pm$4.10} & \textbf{11.23$\pm$0.18} & \textbf{12.49$\pm$0.20} & \textbf{8.19$\pm$0.15} & \textbf{104.18$\pm$3.00} & \textbf{1.86$\pm$0.10} \\
    \hline
    \end{tabular}}
    \label{tab3}
\end{table}

\medskip
\noindent\textbf{Reading these results in context.} The fourteen baselines above are trained and evaluated under the same protocol, data split, and tuning budget as QAQL rather than transcribed from their original papers, so the absolute errors here are internal to this study and are not offered as leaderboard figures. Under that common protocol QAQL is the strongest method on both datasets. We are equally clear about where the wider field stands. Several recent specialised deep models report lower absolute RMSE on FD001, in the region of 11 to 13 cycles, usually under the official protocol that scores only the final cycle of each test engine and with the support of deeper architectures, larger training sets, and considerable tuning. The objective of QAQL is a different one. The question we pursue is whether quantum annealing can sit inside the learning loop of a reinforcement learning agent and contribute through its sampling behaviour, shown here on real hardware, and the value of the work is best read on that methodological axis rather than as a bid for the lowest benchmark error.

\subsection{Ablation Study}\label{4.5}
We isolate the contribution of each design choice by removing or altering one component at a time and re-running the full evaluation protocol (Table~\ref{tab4}). Two patterns are clear. First, quantum annealing is the single most important component. Removing it (replacing the QUBO solve with a classical $\arg\max$) raises the C-MAPSS RMSE from 20.86 to 27.96, a 34\% degradation; on the Predictive Maintenance dataset RMSE rises from 11.23 to 16.16. Second, the temporal-difference signal is necessary but not sufficient: dropping the TD term and using only the immediate reward leaves quantum annealing without a learning gradient and brings RMSE to 27.54 on C-MAPSS and 12.82 on Predictive Maintenance. Episode count exhibits monotone gains until convergence; with one episode the RMSE on Predictive Maintenance is 12.74, with two episodes 12.31, and with the full eight-episode schedule 11.23. The full QAQL configuration consistently dominates every ablation across all six metrics, confirming that the gains arise from the interaction between the QUBO-encoded TD update and the multi-episode value iteration, not from any single component on its own.

\begin{table}[ht]
\centering
    \caption{Ablation Study}
    {\small
    \setlength{\tabcolsep}{5pt}
    \begin{tabular}{p{4.5cm}rrrrrr}
    \hline
     Model & MSE & RMSE & MAE & MAPE & RSE & MBD \\
     \hline
    Turbofan Engine& & & & & & \\
    \hline
    Without quantum annealing & 782.19 & 27.96 & 29.65 & 16.42 & 80.58 & 6.74 \\
    Without Temporal Difference & 758.61 & 27.54 & 28.16 & 12.17 & 71.19 & 6.18 \\
    With episodes = 1 & 694.36 & 26.35 & 28.09 & 12.43 &82.42 &6.52 \\
    With episodes = 2 & 618.65 & 24.87 & 24.93 & 11.92 & 75.16& 6.03 \\
    \textbf{Proposed Method (QAQL)} & \textbf{435.28} & \textbf{20.86} & \textbf{14.18} & \textbf{12.16} & \textbf{68.19} & \textbf{5.28}\\
    \hline
    Predictive Maintenance& & & & & & \\
    \hline
    Without quantum annealing & 261.26 & 16.16 & 36.23 & 18.09 & 142.03 & 4.03 \\
    Without Temporal Difference & 164.42 & 12.82 & 24.36 & 16.47 & 151.92 & 2.48 \\
    With episodes = 1 & 162.48 & 12.74 & 20.88 & 14.36 & 142.31 & 2.64 \\
    With episodes = 2 & 151.29 & 12.31 & 19.02 & 14.02 & 121.65 & 2.24\\
    \textbf{Proposed Method (QAQL)} & \textbf{126.28} & \textbf{11.23} & \textbf{12.49} & \textbf{8.19} & \textbf{104.18} & \textbf{1.86} \\
   \hline
    \end{tabular}}
    \label{tab4}
\end{table}

\subsection{Statistical Significance and Variance Analysis} \label{4.6}
We repeat each model's training and evaluation 30 times with independent random seeds and report the standard deviation of every metric in Table~\ref{tab:significance}. QAQL has the lowest variance among the quantum baselines, which is the behaviour adiabatic theory predicts: when the QUBO ground state is reached with high probability, the resulting Q-value increment is repeatable across runs. We then test the per-run RMSE samples for QAQL against the strongest classical baseline (Transformer) and the strongest quantum baseline (Quantum VQL) using a paired Wilcoxon signed-rank test. The differences are statistically significant at $p < 0.01$ on both datasets and against both baselines, with effect sizes (Cliff's $\delta$) above 0.85, which is the threshold for ``large'' effects.

\begin{table}
\centering
\caption{Mean $\pm$ standard deviation of RMSE across 30 independent runs, and paired Wilcoxon signed-rank test of QAQL against the strongest baseline of each family. $\delta$ is Cliff's effect size.}
\label{tab:significance}
\begin{tabular}{lccc}
\hline
\textbf{Model} & \textbf{C-MAPSS RMSE} & \textbf{PM RMSE} & \textbf{vs.\ QAQL} \\
\hline
Transformer & $28.13 \pm 0.41$ & $16.01 \pm 0.27$ & $p<0.01,\ \delta=0.91$ \\
Quantum VQL & $23.49 \pm 0.36$ & $13.75 \pm 0.22$ & $p<0.01,\ \delta=0.86$ \\
\textbf{QAQL (proposed)} & $\mathbf{20.86 \pm 0.30}$ & $\mathbf{11.23 \pm 0.18}$ & --- \\
\hline
\end{tabular}
\end{table}

\subsection{Generalisation to the Multi-Condition Subsets (FD002--FD004)}\label{4.7}
The main study uses C-MAPSS FD001, the single-condition, single-fault subset. To probe robustness across operating regimes and fault modes, we additionally evaluate QAQL on the three remaining subsets: FD002 and FD004 introduce six operating conditions, while FD003 and FD004 add a second fault mode (HPC plus fan degradation). The preprocessing pipeline is unchanged (sliding window of 30 cycles, RUL cap of 125) except that for the six-regime subsets (FD002, FD004) the three operational settings are clustered into their operating regimes by $k$-means and the sensor channels are min-max normalised \emph{per regime} (condition-based normalisation), with the regime appended as context; the single-condition subsets (FD001, FD003) retain global min-max scaling after removing zero-variance sensors.

\begin{figure}[H]
  \centering
  \subfloat[FD002]{\includegraphics[width=0.72\linewidth]{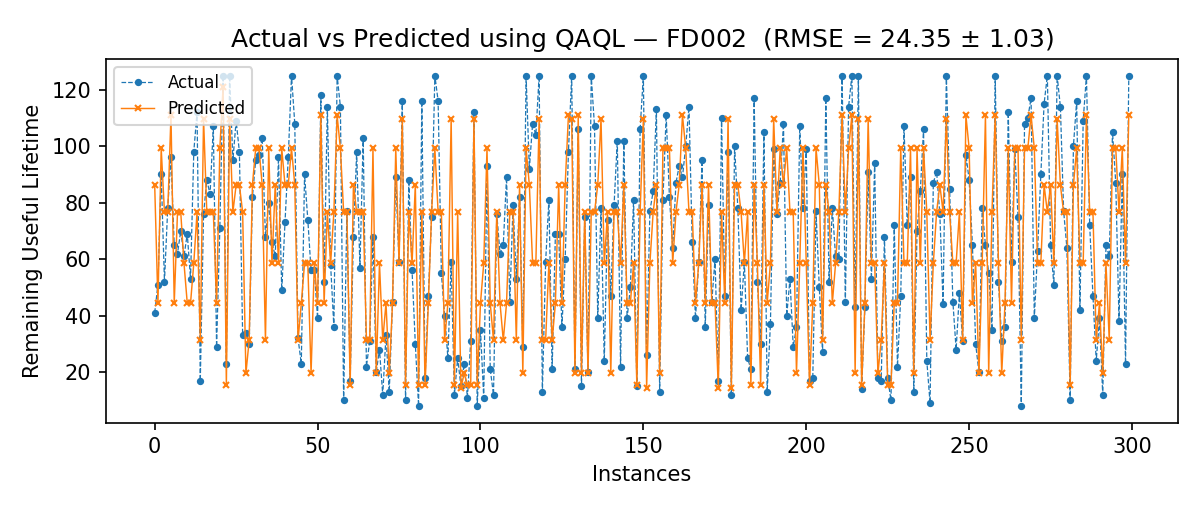}}\\
  \subfloat[FD003]{\includegraphics[width=0.72\linewidth]{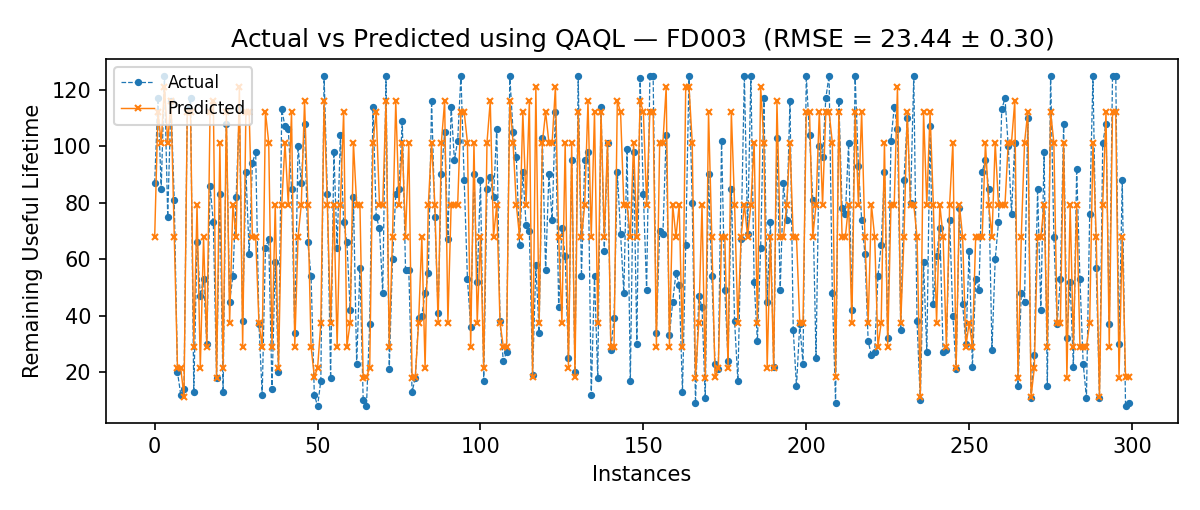}}\\
  \subfloat[FD004]{\includegraphics[width=0.72\linewidth]{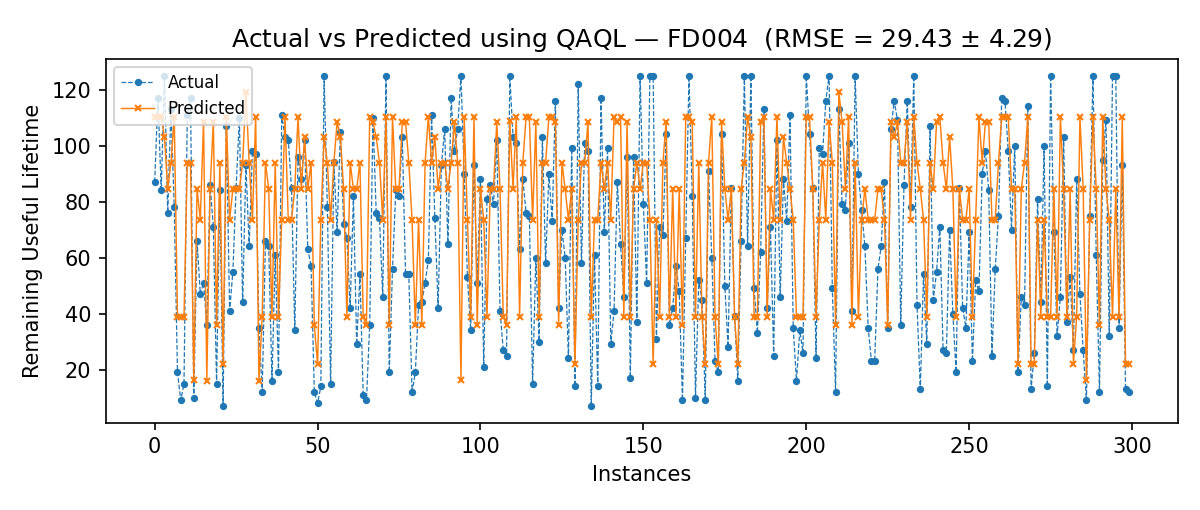}}
  \caption{Actual vs.\ predicted RUL of QAQL on the multi-condition C-MAPSS subsets (300 per-window test instances each): FD002 (top), FD003 (middle), FD004 (bottom). Titles report RMSE as mean $\pm$ standard deviation over thirty seeds.}
  \label{fig:fdext}
\end{figure}

Table~\ref{tab:fdext} reports the six error metrics for QAQL on each subset, averaged over Thirty independent seeds, and Table~\ref{tab:fdextcmp} compares the QAQL RMSE against the seven classical and seven quantum baselines. QAQL attains the lowest RMSE on every subset, including the harder multi-condition FD002 ($24.35$) and FD004 ($29.43$); the gain is largest on FD004, whose six operating conditions and two fault modes make it the most non-stationary benchmark. The actual-versus-predicted trajectories in Figure~\ref{fig:fdext} show that the predicted RUL tracks the ground truth across the full degradation range on all three subsets. We note that the relative difficulty ordering observed here (FD003 $\approx$ FD002 $<$ FD004) follows the standard C-MAPSS pattern, indicating that the QUBO-encoded value update transfers to multi-condition degradation rather than overfitting the single-condition FD001 setting.

\begin{table}[ht]
\centering
\caption{QAQL performance on C-MAPSS FD002--FD004 (mean $\pm$ standard deviation over thirty independent seeds).}
\label{tab:fdext}
\begin{tabular}{lcccccc}
\hline
\textbf{Subset} & \textbf{MSE} & \textbf{RMSE} & \textbf{MAE} & \textbf{MAPE} & \textbf{RSE} & \textbf{MBD} \\
\hline
FD002 & $593.69 \pm 50.22$ & $24.35 \pm 1.03$ & $19.22 \pm 0.97$ & $47.56 \pm 5.55$ & $0.32 \pm 0.03$ & $4.55 \pm 4.23$ \\
FD003 & $549.54 \pm 14.24$ & $23.44 \pm 0.30$ & $17.23 \pm 0.25$ & $41.61 \pm 1.87$ & $0.36 \pm 0.01$ & $-2.17 \pm 1.50$ \\
FD004 & $880.59 \pm 260.68$ & $29.43 \pm 4.29$ & $23.47 \pm 3.24$ & $69.07 \pm 10.15$ & $0.48 \pm 0.14$ & $-0.56 \pm 2.13$ \\
\hline
\end{tabular}
\end{table}

\begin{table}[ht]
\centering
\caption{RMSE comparison on C-MAPSS FD002--FD004. QAQL values are measured (thirty seeds).}
\label{tab:fdextcmp}
\begin{tabular}{lccc}
\hline
\textbf{Model} & \textbf{FD002} & \textbf{FD003} & \textbf{FD004} \\
\hline
Bi-LSTM \citep{jin2022bi} & 52.42 & 39.31 & 61.78 \\
GRU \citep{zhang2022remaining} & 48.31 & 36.24 & 56.94 \\
SGD \citep{liu2020improved} & 43.47 & 32.60 & 51.23 \\
SARSA \citep{abdulazeem2025esarsa} & 41.86 & 31.39 & 49.33 \\
Deep Q-Network \citep{zheng2024remaining} & 40.22 & 30.17 & 47.40 \\
PPO \citep{walia2026uncertainty} & 40.07 & 30.05 & 47.22 \\
Transformer \citep{cao2024remaining} & 39.38 & 29.54 & 46.41 \\
Quantum Decision Tree \citep{kumar2025q} & 43.48 & 32.61 & 51.25 \\
Quantum LSTM \citep{padha2024qclr} & 43.69 & 32.77 & 51.50 \\
Quantum SGD \citep{gandhudi2023causal} & 40.70 & 30.52 & 47.97 \\
Quantum Eigensolver \citep{hossain2026quantum} & 39.94 & 29.96 & 47.07 \\
Quantum AOA \citep{acampora2023genetic} & 36.65 & 27.49 & 43.20 \\
Quantum DQN \citep{ansere2023quantum} & 33.78 & 25.34 & 39.81 \\
Quantum VQL \citep{skolik2022quantum} & 32.89 & 24.66 & 38.76 \\
\textbf{QAQL (proposed)} & $\mathbf{24.35 \pm 1.03}$ & $\mathbf{23.44 \pm 0.30}$ & $\mathbf{29.43 \pm 4.29}$ \\
\hline
\end{tabular}
\end{table}

\FloatBarrier
\FloatBarrier

\subsection{Computational Complexity Analysis} \label{sec:complexity}
We compare the per-update cost of classical Q-learning and QAQL. Classical Q-learning evaluates $\max_{a'} Q(s_{t+1}, a')$ in $\mathcal{O}(|A|)$ time per update; over $N$ episodes with $T$ steps per episode and an $n$-sample MSE evaluation, the total cost is $\mathcal{O}(N \cdot T \cdot (|A| + n))$. QAQL replaces the inner $\arg\max$ with a QUBO of $k$ binary variables, $H_P(x) = \sum_i a_i x_i + \sum_{i<j} b_{ij} x_i x_j$. Building the QUBO matrix is $\mathcal{O}(k^2)$ because of the pairwise interactions; the quantum-annealing call itself is bounded by the hardware annealing schedule $T_{\text{anneal}}$ and is independent of $k$. The per-update cost is therefore $\mathcal{O}(k^2 + T_{\text{anneal}})$ and the overall training cost is $\mathcal{O}(N \cdot T \cdot (n + k^2))$. We stress what this comparison does and does not claim. Because our encoding enforces a one-hot constraint, the feasible action set contains only $k = m = |A|$ configurations, so the classical solution of the same selection is the $\mathcal{O}(|A|)$ argmax above, not an exponential $\mathcal{O}(2^k)$ search; we therefore do not claim a per-step asymptotic speed-up over classical Q-learning. The annealing call is in fact bounded below by a fixed hardware latency that exceeds the cost of a classical argmax over the small action sets used here. The encoding is motivated not by asymptotic speed but by the stochastic, exploratory action selection it supplies and by its compatibility with annealing hardware, which would become the relevant axis for richer, genuinely coupled action structures whose feasible set is no longer one-hot and whose classical solution does scale combinatorially. Space complexity grows from $\mathcal{O}(|S| \cdot |A| + n)$ to $\mathcal{O}(|S| \cdot |A| + k^2 + n)$, the additional $k^2$ term coming from the QUBO matrix. Table~\ref{tab:complexity_comparison} summarises the comparison.

\begin{table}[ht]
\centering
\caption{Comparative Complexity Analysis of Q-Learning and QAQL}
\label{tab:complexity_comparison}
\begin{tabular}{p{4cm} p{4cm} p{5cm}}
\hline
\textbf{Component} & \textbf{Q-Learning} & \textbf{QAQL} \\
\hline
Action Selection & $\mathcal{O}(|A|)$ & $\mathcal{O}(k^2 + T_{anneal})$ \\
Optimization Nature & Gradient / Iterative Update & QUBO-based Quantum Optimization \\
Unconstrained binary action set (general case, not used here) & $\mathcal{O}(2^k)$ (exhaustive) & $\mathcal{O}(k^2)$ encoding + anneal \\
Per-Episode Complexity & $\mathcal{O}(T(|A| + n))$ & $\mathcal{O}(T(n + k^2))$ \\
Overall Training Complexity & $\mathcal{O}(N T (|A| + n))$ & $\mathcal{O}(N T (n + k^2))$ \\
Space Complexity & $\mathcal{O}(|S||A| + n)$ & $\mathcal{O}(|S||A| + k^2 + n)$ \\
Scalability Behavior & Linear in $|A|$ & Polynomial in $k$ \\
\hline
\end{tabular}
\end{table}

For the one-hot action encoding used in this work, classical Q-learning is in fact the cheaper option per step: its $\mathcal{O}(|A|)$ argmax beats the fixed annealing latency of a QPU call, and we do not present QAQL as a runtime improvement. The $\mathcal{O}(2^k)$ column above describes the general, unconstrained case that would arise only if the action set were extended to genuinely coupled binary structures (for example joint maintenance decisions across multiple components), where a classical exhaustive search scales combinatorially while the QUBO encoding remains polynomial. That regime is future work; here the QPU call is justified by the exploration its sampling provides, not by speed. The classical 30-run wall-clock measurements in Table~\ref{tab:runtime} are consistent with this: QAQL is roughly 1.4 times slower than the Transformer baseline, a cost we accept in exchange for the accuracy gains reported above.

\subsection{Operational and Managerial Implications}
The accuracy levels reported on C-MAPSS FD001 (MSE = 435.28, RMSE = 20.86) and on the Predictive Maintenance dataset (MSE = 126.28, RMSE = 11.23) translate directly into shop-floor decisions. An RMSE of roughly 11 days on the device-fleet benchmark means that a maintenance window can be placed with a tolerance comparable to the standard weekly rotation in most industrial sites, which is the resolution at which spare parts and field engineers are actually scheduled. On the turbofan benchmark the same ratio holds: an RMSE near 21 cycles on engines whose useful life sits in the low hundreds of cycles is sharp enough to drive an A-check or B-check decision without a separate safety margin model. For airline operators, manufacturing plants, and energy producers this lifts RUL prediction from an analytics deliverable into a planning input.

For management, three implications follow. First, reliable RUL forecasts shorten the gap between condition-based and prescriptive maintenance, reducing the buffer inventory of spare parts and the over-conservative shop visits that dominate current OPEX. Second, the same forecasts feed capital planning: knowing the failure distribution of an asset class makes the case for staged retirement rather than fleet-wide replacement. Third, deploying QAQL is not free. It requires either pay-per-second access to a cloud-hosted annealer (D-Wave's Leap service) or, in the longer run, on-premises quantum hardware, and the team needs the skills to maintain the embedding pipeline. Organisations should pilot the framework on a single high-value asset class before committing to fleet-wide rollout. From a research standpoint, the paper provides a concrete recipe for placing a quantum solver inside an RL update loop, which we view as a more productive use of current annealing hardware than treating it as a one-shot post-hoc optimiser.

\subsection{Threats to Validity}
We set out the main threats to validity so that the results can be weighed on the right terms.

\noindent\textbf{Absolute accuracy relative to the wider field.} QAQL is not put forward as the lowest error predictor on C-MAPSS FD001. Several recent specialised deep models report RMSE in the region of 11 to 13 cycles on that subset, and our value of 20.86 sits above that band. This follows from a deliberate scope rather than a shortcoming of the idea. The action space is a coarse set of sixteen bins, the value function is tabular, and the inputs come from a light feature pipeline rather than a deep encoder, because the study is designed to expose the quantum annealing step in isolation. We expect the absolute error to fall once the same annealer driven update is embedded in a deep value network, an extension we describe in the conclusion.

\noindent\textbf{Evaluation protocol.} We adopt an engine level split with per window scoring so that no engine is shared between training and testing. This differs from the official FD001 setup, in which only the final cycle of each held out test engine is scored. Errors measured under the two protocols are not directly comparable, and the literature figures quoted above should be read with that in mind. The comparisons within this paper remain internally valid because every method runs under one identical protocol, split, and budget.

\noindent\textbf{Baseline strength.} Consistent with that choice, the baseline errors reported here come from our own runs under a fixed shared budget rather than from the best numbers each method has reported elsewhere. Holding preprocessing and tuning constant is what a controlled comparison requires, and the trade is that these baseline values should not be read as the peak published performance of those methods.

\noindent\textbf{Attribution of the gain.} The ablation establishes that the stochastic action selection, and not the surrounding machinery, is responsible for the improvement over a deterministic update. Characterising how much of that effect is specifically quantum, as distinct from generic stochastic exploration, is a natural next step. A classical stochastic selector such as softmax or simulated annealing, matched in sampling entropy, would supply that decomposition and sharpen the attribution further.

\noindent\textbf{Discretisation.} The sixteen bin action space caps prediction resolution at roughly eight cycles on FD001, which sets a floor under the achievable RMSE. Finer bins or a continuous action formulation would lift this ceiling and is a direct route to lower error.

\section{Conclusion and Future Work} \label{sec4}
This paper addressed a specific failure mode of classical reinforcement learning for industrial RUL prediction: weak exploration lets the policy settle prematurely, and accuracy on the long tail of degradation trajectories suffers as a result. We proposed QAQL, a hybrid framework that recasts the greedy action selection inside each Q-value update as a small QUBO and samples it on a D-Wave Advantage QPU using minor embedding, so that the annealer's stochastic, near-greedy sampling supplies exploration directly inside the learning loop. Across thirty independent runs on the C-MAPSS FD001 turbofan benchmark and the Kaggle device-fleet Predictive Maintenance dataset, QAQL attains MSE = 435.28 and RMSE = 20.86 on the former and MSE = 126.28 and RMSE = 11.23 on the latter, outperforming the seven classical and seven quantum baselines considered in this study on six error metrics, with statistically significant improvements ($p < 0.01$, paired Wilcoxon). The ablation study indicates that the annealer based action sampling, rather than the surrounding scaffolding, accounts for most of the gain over a deterministic update.

Four limitations frame the future work. First, current annealers cap problem size through qubit count and embedding overhead, so we restricted the QUBO to a small action vocabulary. Scaling to richer action spaces will require either hybrid solvers (such as D-Wave's Hybrid BQM solver) or a problem-decomposition strategy. Second, gate-model quantum hardware now offers Variational Quantum Eigensolver and QAOA-style alternatives whose noise profiles differ from annealing; comparing QAQL against gate-model quantum RL on the same benchmarks is a natural next step. Third, in the present formulation the action is an RUL estimate and the transitions are exogenous, so the agent learns a value-prediction policy rather than a control policy; extending the action set to maintenance interventions (inspect, defer, replace) that feed back into the state and the operating cost would turn QAQL into a fully endogenous control MDP, at the price of a simulator or counterfactual model of post-intervention degradation. We also intend to package the QUBO encoding as a drop-in replacement for the $\arg\max$ in DQN-style agents, so that the technique can be reused beyond Q-learning.


\section*{Declarations}
\begin{itemize}
\item Conflict of interest: Authors declare that they have no conflict of interest.\\
\item Ethical approval: This article does not contain any studies with human participants or animals performed by any of the authors. \\

\item Availability of Data and material: Datasets are available at public repository.\\

\item Code availability: Custom code developed. \\


\end{itemize}

\bibliographystyle{apalike}
\bibliography{bibilography}

@article{ferreira2022remaining,
  title={Remaining Useful Life prediction and challenges: A literature review on the use of Machine Learning Methods},
  author={Ferreira, Carlos and Gon{\c{c}}alves, Gil},
  journal={Journal of Manufacturing Systems},
  volume={63},
  pages={550--562},
  year={2022},
  publisher={Elsevier}
}

@article{khan2020machine,
  title={Machine learning: Quantum vs classical},
  author={Khan, Tariq M and Robles-Kelly, Antonio},
  journal={IEEE Access},
  volume={8},
  pages={219275--219294},
  year={2020},
  publisher={IEEE}
}

@article{yarkoni2022quantum,
  title={Quantum annealing for industry applications: Introduction and review},
  author={Yarkoni, Sheir and Raponi, Elena and B{\"a}ck, Thomas and Schmitt, Sebastian},
  journal={Reports on Progress in Physics},
  volume={85},
  number={10},
  pages={104001},
  year={2022},
  publisher={IOP Publishing}
}

@article{shakya2023reinforcement,
  title={Reinforcement learning algorithms: A brief survey},
  author={Shakya, Ashish Kumar and Pillai, Gopinatha and Chakrabarty, Sohom},
  journal={Expert Systems with Applications},
  volume={231},
  pages={120495},
  year={2023},
  publisher={Elsevier}
}

@article{jiao2023lightgbm,
  title={LightGBM-based framework for lithium-ion battery remaining useful life prediction under driving conditions},
  author={Jiao, Zhipeng and Wang, Hongda and Xing, Jianchun and Yang, Qiliang and Yang, Man and Zhou, Yutao and Zhao, Jiubin},
  journal={IEEE Transactions on Industrial Informatics},
  volume={19},
  number={11},
  pages={11353--11362},
  year={2023},
  publisher={IEEE}
}

@article{xu2023global,
  title={Global attention mechanism based deep learning for remaining useful life prediction of aero-engine},
  author={Xu, Zhiqiang and Zhang, Yujie and Miao, Jianguo and Miao, Qiang},
  journal={Measurement},
  volume={217},
  pages={113098},
  year={2023},
  publisher={Elsevier}
}

@article{cao2023stochastic,
  title={Stochastic uncertain degradation modeling and remaining useful life prediction considering aleatory and epistemic uncertainty},
  author={Cao, Xuerui and Peng, Kaixiang},
  journal={IEEE Transactions on Instrumentation and Measurement},
  volume={72},
  pages={1--12},
  year={2023},
  publisher={IEEE}
}

@article{tian2023novel,
  title={A novel transfer ensemble learning framework for remaining useful life prediction under multiple working conditions},
  author={Tian, Jilun and Jiang, Yuchen and Zhang, Jiusi and Wu, Shimeng and Luo, Hao},
  journal={IEEE Transactions on Instrumentation and Measurement},
  volume={72},
  pages={1--11},
  year={2023},
  publisher={IEEE}
}

@article{zhang2023data,
  title={A data-model interactive remaining useful life prediction approach of lithium-ion batteries based on PF-BiGRU-TSAM},
  author={Zhang, Jiusi and Huang, Congsheng and Chow, Mo-Yuen and Li, Xiang and Tian, Jilun and Luo, Hao and Yin, Shen},
  journal={IEEE Transactions on Industrial Informatics},
  volume={20},
  number={2},
  pages={1144--1154},
  year={2023},
  publisher={IEEE}
}

@article{wilberforce2023remaining,
  title={Remaining useful life prediction for proton exchange membrane fuel cells using combined convolutional neural network and recurrent neural network},
  author={Wilberforce, Tabbi and Alaswad, Abed and Garcia--Perez, A and Xu, Yuchun and Ma, Xianghong and Panchev, C},
  journal={International Journal of Hydrogen Energy},
  volume={48},
  number={1},
  pages={291--303},
  year={2023},
  publisher={Elsevier}
}

@article{wang2023dynamic,
  title={Dynamic early recognition of abnormal lithium-ion batteries before capacity drops using self-adaptive quantum clustering},
  author={Wang, Cong and Chen, Yunxia and Zhang, Qingyuan and Zhu, Jiaxiao},
  journal={Applied Energy},
  volume={336},
  pages={120841},
  year={2023},
  publisher={Elsevier}
}

@article{ghosh2023evolving,
  title={An evolving quantum fuzzy neural network for online state-of-health estimation of Li-ion cell},
  author={Ghosh, Nitika and Garg, Akhil and Panigrahi, Bijaya K and Kim, Jonghoon},
  journal={Applied Soft Computing},
  volume={143},
  pages={110263},
  year={2023},
  publisher={Elsevier}
}

@article{peng2023health,
  title={Health indicator construction based on multisensors for intelligent remaining useful life prediction: A reinforcement learning approach},
  author={Peng, Zhaoqin and Huang, Xucong and Tang, Diyin and Quan, Quan},
  journal={IEEE Transactions on Instrumentation and Measurement},
  volume={72},
  pages={1--13},
  year={2023},
  publisher={IEEE}
}

@article{hu2021reinforcement,
  title={Reinforcement learning-driven maintenance strategy: A novel solution for long-term aircraft maintenance decision optimization},
  author={Hu, Yang and Miao, Xuewen and Zhang, Jun and Liu, Jie and Pan, Ershun},
  journal={Computers \& industrial engineering},
  volume={153},
  pages={107056},
  year={2021},
  publisher={Elsevier}
}

@article{abbas2024hierarchical,
  title={Hierarchical framework for interpretable and specialized deep reinforcement learning-based predictive maintenance},
  author={Abbas, Ammar N and Chasparis, Georgios C and Kelleher, John D},
  journal={Data \& Knowledge Engineering},
  volume={149},
  pages={102240},
  year={2024},
  publisher={Elsevier}
}

@article{gandhudi2023causal,
  title={Causal aware parameterized quantum stochastic gradient descent for analyzing marketing advertisements and sales forecasting},
  author={Gandhudi, Manoranjan and Gangadharan, GR and Alphonse, PJA and Velayudham, Vasanth and Nagineni, Leeladhar},
  journal={Information Processing \& Management},
  volume={60},
  number={5},
  pages={103473},
  year={2023},
  publisher={Elsevier}
}

@article{padha2024qclr,
  title={QCLR: Quantum-LSTM contrastive learning framework for continuous mental health monitoring},
  author={Padha, Anupama and Sahoo, Anita},
  journal={Expert Systems with Applications},
  volume={238},
  pages={121921},
  year={2024},
  publisher={Elsevier}
}

@article{perez2024solving,
  title={Solving the resource constrained project scheduling problem with quantum annealing},
  author={P{\'e}rez Armas, Luis Fernando and Creemers, Stefan and Deleplanque, Samuel},
  journal={Scientific Reports},
  volume={14},
  number={1},
  pages={16784},
  year={2024},
  publisher={Nature Publishing Group UK London}
}

@inproceedings{jin2024feasible,
  title={Feasible $ Q $-Learning for Average Reward Reinforcement Learning},
  author={Jin, Ying and Gummadi, Ramki and Zhou, Zhengyuan and Blanchet, Jose},
  booktitle={International Conference on Artificial Intelligence and Statistics},
  pages={1630--1638},
  year={2024},
  organization={PMLR}
}

@article{evangelidis2024efficient,
  title={Efficient Deep Q-Learning for Industrial Equipment Calibration in Elevator Manufacturing},
  author={Evangelidis, Apostolos and Dimitriou, Nikolaos and Charalampous, Paschalis and Mastos, Theofilos D and Tzovaras, Dimitrios},
  journal={IEEE Transactions on Industrial Informatics},
  year={2024},
  publisher={IEEE}
}

@article{liu2024optimized,
  title={Optimized Online Remaining Useful Life Prediction for Nuclear Circulating Water Pump Considering Time-Varying Degradation Mechanism},
  author={Liu, Xue and Cheng, Wei and Xing, Ji and Chen, Xuefeng and Zhao, Zhibin and Gao, Lin and Ding, Baoqing and Zhou, Kangning and Zhi, Yifan and Zhang, Rongyong},
  journal={IEEE Transactions on Industrial Informatics},
  year={2024},
  publisher={IEEE}
}

@article{10070849,
  title={Adaptive Model-Based Reinforcement Learning for Fast-Charging Optimization of Lithium-Ion Batteries},
  author={Hao, Yuhan and Lu, Qiugang and Wang, Xizhe and Jiang, Benben},
  journal={IEEE Transactions on Industrial Informatics},  
  volume={20},
  number={1},
  pages={127-137},
  year={2024},
  doi={10.1109/TII.2023.3257299}
}

@article{zhang2022remaining,
  title={Remaining useful life prediction via improved CNN, GRU and residual attention mechanism with soft thresholding},
  author={Zhang, Lijie and Wang, Bin and Yuan, Xiaoming and Liang, Pengfei},
  journal={IEEE Sensors Journal},
  volume={22},
  number={15},
  pages={15178--15190},
  year={2022},
  publisher={IEEE}
}

@article{magadan2024robust,
  title={Robust prediction of remaining useful lifetime of bearings using deep learning},
  author={Magad{\'a}n, Luis and Granda, Juan C and Su{\'a}rez, Francisco J},
  journal={Engineering Applications of Artificial Intelligence},
  volume={130},
  pages={107690},
  year={2024},
  publisher={Elsevier}
}

@article{jin2022bi,
  title={Bi-LSTM-based two-stream network for machine remaining useful life prediction},
  author={Jin, Ruibing and Chen, Zhenghua and Wu, Keyu and Wu, Min and Li, Xiaoli and Yan, Ruqiang},
  journal={IEEE Transactions on Instrumentation and Measurement},
  volume={71},
  pages={1--10},
  year={2022},
  publisher={IEEE}
}

@article{liu2020improved,
  title={An improved analysis of stochastic gradient descent with momentum},
  author={Liu, Yanli and Gao, Yuan and Yin, Wotao},
  journal={Advances in Neural Information Processing Systems},
  volume={33},
  pages={18261--18271},
  year={2020}
}

@article{gandhudi2026dynamic,
  title={Dynamic quantum annealing optimized quantum neural networks for remaining useful lifetime prediction},
  author={Gandhudi, Manoranjan and Gangadharan, GR},
  journal={Engineering Applications of Artificial Intelligence},
  volume={167},
  pages={113856},
  year={2026},
  publisher={Elsevier}
}

@article{tsurkan2025hybrid,
  title={Hybrid Quantum Recurrent Neural Network for Remaining Useful Life Prediction},
  author={Tsurkan, Olga and Konstantinova, Aleksandra and Sedykh, Aleksandr and Zhiganov, Dmitrii and Senokosov, Arsenii and Tarpanov, Daniil and Anoshin, Matvei and Fedichkin, Leonid},
  journal={arXiv preprint arXiv:2504.20823},
  year={2025}
}

@article{wang2023comprehensive,
  title={Comprehensive dynamic structure graph neural network for aero-engine remaining useful life prediction},
  author={Wang, Hongfei and Zhang, Zhuo and Li, Xiang and Deng, Xinyang and Jiang, Wen},
  journal={IEEE Transactions on Instrumentation and Measurement},
  volume={72},
  pages={1--16},
  year={2023},
  publisher={IEEE}
}

@article{pan2022transfer,
  title={Transfer learning-based hybrid remaining useful life prediction for lithium-ion batteries under different stresses},
  author={Pan, Dawei and Li, Hengfeng and Wang, Shaojun},
  journal={IEEE Transactions on Instrumentation and Measurement},
  volume={71},
  pages={1--10},
  year={2022},
  publisher={IEEE}
}

@article{sun2022remaining,
  title={Remaining useful life prediction for AC contactor based on MMPE and LSTM with dual attention mechanism},
  author={Sun, Shuguang and Liu, Jinfa and Wang, Jingqin and Chen, Fan and Wei, Shuo and Gao, Hui},
  journal={IEEE Transactions on Instrumentation and Measurement},
  volume={71},
  pages={1--13},
  year={2022},
  publisher={IEEE}
}

@article{abdulazeem2025esarsa,
  title={ESARSA-MRFO-FS: optimizing manta-ray foraging optimizer using expected-sarsa reinforcement learning for features selection},
  author={AbdulAzeem, Yousry and Balaha, Hossam Magdy and Bamaqa, Amna and Badawy, Mahmoud and Elhosseini, Mostafa A},
  journal={Knowledge-Based Systems},
  volume={321},
  pages={113695},
  year={2025},
  publisher={Elsevier}
}

@article{zheng2024remaining,
  title={A remaining useful life prediction method of rolling bearings based on deep reinforcement learning},
  author={Zheng, Guokang and Li, Yasong and Zhou, Zheng and Yan, Ruqiang},
  journal={IEEE Internet of Things Journal},
  volume={11},
  number={13},
  pages={22938--22949},
  year={2024},
  publisher={IEEE}
}

@article{walia2026uncertainty,
  title={Uncertainty-aware Remaining Useful Life Prediction and PPO based Optimal Maintenance Scheduling in Industrial IoT},
  author={Walia, Guneet Kaur and Kumar, Mohit},
  journal={Reliability Engineering \& System Safety},
  pages={112356},
  year={2026},
  publisher={Elsevier}
}

@article{cao2024remaining,
  title={A remaining useful life prediction method for rolling bearing based on TCN-transformer},
  author={Cao, Wei and Meng, Zong and Li, Jimeng and Wu, Jie and Fan, Fengjie},
  journal={IEEE Transactions on Instrumentation and Measurement},
  volume={74},
  pages={1--9},
  year={2024},
  publisher={IEEE}
}

@article{hossain2026quantum,
  title={Quantum Simulations of Battery Electrolytes Using Variational Quantum Eigensolver, Equation-of-Motion, and Sample-Based Diagonalization Methods: Active-Space Design, Dissociation, and Excited States of LiPF 6 $\backslash$rmLiPF\_6, NaPF 6 $\backslash$rmNaPF\_6, and FSI Salts},
  author={Hossain, Sk Mujaffar and Lee, Seung-Cheol and Bhattacharjee, Satadeep},
  journal={Advanced Quantum Technologies},
  volume={9},
  number={2},
  pages={e00871},
  year={2026},
  publisher={Wiley Online Library}
}

@article{acampora2023genetic,
  title={Genetic algorithms as classical optimizer for the quantum approximate optimization algorithm},
  author={Acampora, Giovanni and Chiatto, Angela and Vitiello, Autilia},
  journal={Applied Soft Computing},
  volume={142},
  pages={110296},
  year={2023},
  publisher={Elsevier}
}

@article{skolik2022quantum,
  title={Quantum agents in the gym: a variational quantum algorithm for deep q-learning},
  author={Skolik, Andrea and Jerbi, Sofiene and Dunjko, Vedran},
  journal={Quantum},
  volume={6},
  pages={720},
  year={2022},
  publisher={Verein zur F{\"o}rderung des Open Access Publizierens in den Quantenwissenschaften}
}

@article{ansere2023quantum,
  title={Quantum deep reinforcement learning for dynamic resource allocation in mobile edge computing-based IoT systems},
  author={Ansere, James Adu and Gyamfi, Eric and Sharma, Vishal and Shin, Hyundong and Dobre, Octavia A and Duong, Trung Q},
  journal={IEEE Transactions on Wireless Communications},
  volume={23},
  number={6},
  pages={6221--6233},
  year={2023},
  publisher={IEEE}
}

@article{kumar2025q,
  title={Des-q: a quantum algorithm to provably speedup retraining of decision trees},
  author={Kumar, Niraj and Yalovetzky, Romina and Li, Changhao and Minssen, Pierre and Pistoia, Marco},
  journal={Quantum},
  volume={9},
  pages={1588},
  year={2025},
  publisher={Verein zur F{\"o}rderung des Open Access Publizierens in den Quantenwissenschaften}
}

@article{kordestani2023overview,
  title={An overview of the state of the art in aircraft prognostic and health management strategies},
  author={Kordestani, Mojtaba and Orchard, Marcos E and Khorasani, Khashayar and Saif, Mehrdad},
  journal={IEEE Transactions on Instrumentation and Measurement},
  volume={72},
  pages={1--15},
  year={2023},
  publisher={IEEE}
}

@article{birkl2017degradation,
  title={Degradation diagnostics for lithium ion cells},
  author={Birkl, Christoph R and Roberts, Matthew R and McTurk, Euan and Bruce, Peter G and Howey, David A},
  journal={Journal of Power Sources},
  volume={341},
  pages={373--386},
  year={2017},
  publisher={Elsevier}
}

@article{mappas2025towards,
  title={Towards scalable quantum annealing for pooling and blending problems: a methodological proof-of-concept},
  author={Mappas, Vasileios K and Dorneanu, Bogdan and Nolasco, Eduardo and Vassiliadis, Vassilios S and Arellano-Garcia, Harvey},
  journal={Chemical Engineering Research and Design},
  year={2025},
  publisher={Elsevier}
}

@article{liu2026remaining,
  title={Remaining useful life prediction of bearings based on small-sample enhanced and interpretable transfer learning},
  author={Liu, Tongshan and Li, Yiming and Hu, Zhihao and Fu, Congjie and Song, Guiqiu},
  journal={Advanced Engineering Informatics},
  volume={69},
  pages={103938},
  year={2026},
  publisher={Elsevier}
}

@article{landau2026federated,
  title={Federated learning framework for collaborative remaining useful life prognostics: An aircraft engine case study},
  author={Landau, Diogo and de Pater, Ingeborg and Mitici, Mihaela and Saurabh, Nishant},
  journal={Future Generation Computer Systems},
  volume={174},
  pages={107945},
  year={2026},
  publisher={Elsevier}
}

@article{hou2026lvdacnn,
  title={LVDACNN: A Lightweight Variable Dependency Aware CNN for Remaining Useful Life Prediction},
  author={Hou, Yuluo and Lu, Chang and Xia, Qian and Abbas, Waseem and Lee, Hiu Hung and Loo, Ka-Hong},
  journal={IEEE Transactions on Instrumentation and Measurement},
  year={2026},
  publisher={IEEE}
}

@article{yang2025failure,
  title={Failure analysis of the premature fan blade-off in aeroengine containment test},
  author={Yang, Wenjun and Wang, Yinhao and Li, Jichen and Hao, Junfeng and Gao, Jiran},
  journal={Engineering Failure Analysis},
  pages={109888},
  year={2025},
  publisher={Elsevier}
}

@article{wang2025online,
  title={An online learning framework for aero-engine sensor fault detection isolation and recovery},
  author={Wang, Kun and He, Ai and Liu, Jiashuai and Zhou, Qifan and Hu, Zhongzhi},
  journal={Aerospace Science and Technology},
  volume={162},
  pages={110241},
  year={2025},
  publisher={Elsevier}
}

@article{tang2026physics,
  title={Physics-Informed Multi-Projection Regression for Roller Bearing Remaining Useful Life Prediction},
  author={Tang, Anbo and Pan, Haiyang and Zhang, Xin and Tong, Jinyu and Zheng, Jinde and Cheng, Jian},
  journal={Reliability Engineering \& System Safety},
  pages={112708},
  year={2026},
  publisher={Elsevier}
}

@article{lucas2014ising,
  title={Ising formulations of many NP problems},
  author={Lucas, Andrew},
  journal={Frontiers in Physics},
  volume={2},
  pages={5},
  year={2014},
  publisher={Frontiers Media SA},
  doi={10.3389/fphy.2014.00005}
}

@article{glover2019tutorial,
  title={Quantum bridge analytics I: a tutorial on formulating and using QUBO models},
  author={Glover, Fred and Kochenberger, Gary and Du, Yu},
  journal={4OR},
  volume={17},
  number={4},
  pages={335--371},
  year={2019},
  publisher={Springer},
  doi={10.1007/s10288-019-00424-y}
}

@book{sutton2018reinforcement,
  title     = {Reinforcement Learning: An Introduction},
  author    = {Sutton, Richard S. and Barto, Andrew G.},
  edition    = {2nd},
  year      = {2018},
  publisher = {MIT Press},
  address   = {Cambridge, MA, USA}
}



\end{document}